# Covid-19 Discourse on Twitter: How the Topics, Sentiments, Subjectivity, and Figurative Frames Changed Over Time

Philipp Wicke[1] and Marianna M. Bolognesi[2]*

[1] Creative Language Systems, School of Computer Science, University College Dublin, Dublin, Ireland, [2] Department of Modern Languages, Literatures, and Culture, University Bologna, Bologna, Italy

The words we use to talk about the current epidemiological crisis on social media can inform us on how we are conceptualizing the pandemic and how we are reacting to its development. This paper provides an extensive explorative analysis of how the discourse about Covid-19 reported on Twitter changes through time, focusing on the first wave of this pandemic. Based on an extensive corpus of tweets (produced between 20th March and 1st July 2020) first we show how the topics associated with the development of the pandemic changed through time, using topic modeling. Second, we show how the sentiment polarity of the language used in the tweets changed from a relatively positive valence during the first lockdown, toward a more negative valence in correspondence with the reopening. Third we show how the average subjectivity of the tweets increased linearly and fourth, how the popular and frequently used figurative frame of WAR changed when real riots and fights entered the discourse.

**Keywords: twitter, corpus analysis, covid-19, topic modeling, sentiment analysis, figurative framing**

## INTRODUCTION

Covid-19 was first officially reported by the Chinese authorities as a virus originated in Wuhan city, Hubei province in China, on 31st December 2019. According to official notifications of the World Health Organization (2020), while we revise this manuscript the disease has infected more than 106 million people worldwide, killing more than 2.3 million lives.

The issues related to the development of the global pandemic are challenging and complex, because they carry deep consequences not only in the medical, but also the social, economic, political, and behavioral domains. While the recent release of different types of vaccines suggest that we might be experiencing the last phases of this health crisis, the consequences of such a long-lasting worldwide pandemic will be certainly seen beyond the actual end of the medical emergency and in various aspects of our lives.

Online discourse on Twitter, in this regard, has recently attracted a number of contributions, because the texts (the tweets) found on this platform are considered to be a good proxy for the public opinion and perception related to the pandemic that we are currently experiencing (Bruns and Weller, 2016). It follows that understanding and interpreting such discourse, its evolution over time, and its interdependence with real-world events can help us understand how people conceptualize and react to the global crisis.

In particular, understanding how the topics discussed on Twitter in relation to the pandemic change over time can be crucial for understanding what aspects of the crisis are perceived to be more salient and important for the population (Zhou et al., 2020). In a very recent study Wicke and Bolognesi (2020) analyzed the topics of discussion in a corpus of tweets that covered 2 months (20th March—20th May 2020). In the discussion of their findings, the authors suggest that topics are likely to change over time. Therefore, adding a temporal dynamic to the topic modeling analysis may provide a clearer view of how the pandemic is processed in the minds of the speakers and discussed on Twitter.

Mining the sentiment polarity of tweets through the analysis of words used therein can provide precious information about how social measures such as travel bans, social distancing, and so forth have been taken in by the population during the first wave. By seeing potential changes in the sentiment polarity through time, and interpreting them in relation with major events and governmental decisions issued during the first wave, it may become possible to predict how similar measures are going to affect us now that we are experiencing a new wave.

If tweets that contain language loaded with affective information are likely to express opinions rather than facts, then they therefore tend to be subjective rather than objective. Mining the amount of affective information (positive or negative) associated with the language used in the tweets can shed light on the temporal dynamics of the overall subjectivity of the tweets. In other words, it will be possible to observe the distribution of fact-based vs. opinion-based tweets over time (De Smedt and Daelemans, 2012). This type of analysis can provide an interesting indicator of our eagerness to report, trust, and discuss facts and potential objective information, as opposed to opinions.

Finally, understanding how a specific conceptual framing used in the discourse about Covid on Twitter changes over time can provide a different type of indirect measure of people's attitude toward the pandemic. In particular, previous research has shown that various sensitive topics such as cancer, drugs, crime, and epidemics are typically framed using the pervasive metaphorical frame of WAR (Flusberg et al., 2017; Thibodeau et al., 2017; Wicke and Bolognesi, 2020). In some cases, however, the use of war-related terms to talk about sensitive topics has been proven to have negative effects on the people directly affected by the problem under discussion. For example, using war-related terms to talk about cancer affects patients' general attitude toward their own medical condition (Hendricks et al., 2018). Conversely, the use of alternative, more positive frames, such as JOURNEY or DANCE, can positively affect patients' attitude and general well-being. Since previous work has shown that, generally speaking, the WAR frame is particularly frequent in the discourse about Covid-19 (Wicke and Bolognesi, 2020), we hereby explore how the distribution of the lexical units within this figurative frame change over time, to possibly cover and express topics associated with the new stages of the pandemic, in a temporal perspective.

In line with the variables outlined above, the research questions addressed in this study can be summarized as follows:

1. Which topics are discussed on Twitter in relation to Covid and how do they change over time, with the development of the pandemic?
2. What valence (sentiment polarity) emerges from the tweets about Covid and how does it change over time?
3. How does the subjectivity of the tweets (i.e., opinion-based focus, vs. the objective fact-based focus) change over time?
4. How does the use of the pervasive figurative framing of WAR change over time?

Following the research questions outlined above, we formulated the following hypotheses.

1. TOPICS: The pandemic is in constant development and change. The topics of discussion on Twitter are likely to change accordingly, in concurrence with the most recent events associated with Covid-19. We therefore predict that different topic models, based on different degrees of granularity will capture different events covered by the media and the press, related to Covid-19.
2. SENTIMENT POLARITY: The corpus of tweets on which the current analysis is performed contains mainly data produced by American English users, collected between 20th March (first official day of lockdown in many States) and 1st July 2020. In this period of time the number of active cases increased steadily in the USA, according to the World Health Organization (2020). We therefore expect to find an increase in the negative feelings associated with the tweets, over time.
3. SUBJECTIVITY: Because of the development of the pandemic, and the increase of the daily cases, and of the (possibly) negative feelings emerging from the tweets, we expect the tweets to contain an increasing number of words loaded with affective content. It follows that we expect the tweets to be increasingly opinion-based (loaded with emotion), rather than fact-based (neutral), with the progressing of the epidemic.
4. FRAMING: We do not have a specific hypothesis in mind in relation to this research question, but we expect to observe possible changes in the way in which the WAR frame is used to talk about the virus. In particular, while words such as "fight" and "war" may continue to be frequently used, we might observe new words within this frame becoming common in the Covid discourse. This would suggest that the lexical tools used to frame the Covid discourse have been extended and developed, to confirm the centrality and pervasiveness of the WAR figurative frame.

The remainder of the paper is organized as follows: after a brief overview of related work on these topics, we proceed by addressing each research question in order, explaining methods, results, and discussion of the data related to each analysis. Finally, we take all the results together and provide a final general discussion of our findings.

## THEORETICAL BACKGROUND AND RELATED WORK

The information encoded in the short texts produced by private internet users on Twitter (the tweets) provides useful clues that in some cases can be used by experts. A growing body of research on social media discourse associated with disasters and crises is based on Twitter. Yeo et al. (2018) for instance, reported a study of social media discourse about the 2016 Southern Louisiana flooding in which they used Twitter data to construct a response communication network and show culture-specific characteristics of this discourse. In a more recent study, Yeo et al. tracked topics, sentiments, and patterns on longitudinal Twitter data on the same phenomenon (Yeo et al., 2020). Thanks to this analysis they provided an overview of the long-term crisis recovery with respect to the dominant voices, sentiments, and participants' numbers. The authors highlighted the need for long-term recovery communication, utilizing social media, and supporting local voices after a disaster. A spatiotemporal analysis of the Twitter discourse about Hurricane Matthew has been conducted by Martín et al. (2017). The authors conducted a temporal analysis and tracked disaster-related tweets over a week for different states in the US in order to correlate the distance to the hurricane with Twitter activity. With a fine-grained analysis they were able to observe evacuees and traveling information during the development of this disaster, which allowed them to check evacuation compliance.

In relation to previous epidemics, the linguistic data extracted from Twitter has been correlated with the actual spreading of the virus, showing that the number of tweets discussing flu-symptoms predicted the official statistics about the virus spread such as those published by Centers for Disease Control and Prevention and the Health Protection Agency (Culotta, 2010). Quantitative analyses of linguistic data been conducted during the development of various types of diseases to mine the information that internet users encode in language, while experiencing medical crises such as the dengue fever in Brazil (Gomide et al., 2011), the Zika disease (Wirz et al., 2018; Pruss et al., 2019), the measles outbreak in the Netherlands in 2013 (Mollema et al., 2015), and more recently, the Coronavirus epidemic (Wicke and Bolognesi, 2020).

## Topics

In relation to the spreading of the Zika virus in 2015, Miller and colleagues (Miller et al., 2017) used a combination of natural language processing and machine learning techniques to determine the distribution of topics in relation to four characteristics of Zika: symptoms, transmission, prevention, and treatment. The authors reported the most persistent concerns or misconceptions regarding the Zika virus extracted from the corpus of tweets, and provided a complex map of topics that emerged from the analysis. For example, in relation to the prevention of the virus spreading, they observed the emergence of the following topics: need for control, and prevention of spread, need for money, ways to prevent spread, bill to get funds, and research. In a different study, Pruss et al. (2019) provided a cross-linguistic analysis of the discourse around the Zika virus, based on a corpus of tweets in three different languages (Spanish, Portuguese, and English). Using a multilingual topic model, the authors identified key topics of discussion across the languages and their distribution, demonstrating that the Zika outbreak was discussed differently around the world. Lazard and colleagues, instead, analyzed the topics related to the discourse around the Ebola outbreak, in 2014, and in particular after a case of Ebola was diagnosed on US soil. The authors reported that the main topics of concern for the American public were the symptoms and lifespan of the virus, the disease transfer and contraction, whether it was safe to travel, and how they could protect themselves from the disease. In a parallel study, Tran and Lee (2016) built Ebola-related information propagation models to mine the information encoded in the tweets about Ebola and explored how such information is distributed across the following six topics: 1. Ebola cases in the US, 2. Ebola outbreak in the world, 3. fear and prayer, 4. Ebola spread and warning, 5. jokes, swearing, and disapproval of jokes and 6. impact of Ebola to daily life. The authors found that the second topic had the lowest focus, while the fifth and sixth had the highest. Finally, in a very recent study, Park et al. (2020) propose a topic analysis related to the discourse around Covid on Twitter, analyzing a corpus of Indian, South Korean, Vietnamese, and Iranian tweets in a temporal perspective. The authors report some cultural differences, showing that in Iran and Vietnam, unlike in South Korea, the number of tweets did not correlate with the dates of specific events taking place in these countries, which were used by the authors as baselines. In a temporal analysis they report that the official epidemic phases issued by governments do not match well with the online attention on the epidemic. Nonetheless, the authors compared similarities in major topics across these countries over time and found that in Iran, Vietnam, and India, the peak of the daily tweet trend preceded the peak of the daily confirmed cases. This suggests that mining tweets can help to monitor public attention toward the diffusion of the epidemic.

Finally, Twitter-based studies that use topic modeling techniques or sentiment analysis are starting to appear in relation to the Covid discourse. However, to the best of our knowledge, they appear to use a significantly different methodology. Those works include Sentiment Analysis with Deep Learning Classifiers (Chakraborty et al., 2020; Li et al., 2020), a time-span of much fewer days (Abd-Alrazaq et al., 2020; Xue et al., 2020), and analyses of specific emotions without topic models (Lwin et al., 2020; Mathur et al., 2020). Because topic modeling is an exploratory, bottom-up, data-driven technique of data mining, we believe that a broader and more explorative approach, that takes into account multiple topic modeling solutions and a longer time span, may provide better insights on the themes discussed by Twitter users over time.

## Sentiment Polarity

Many of these linguistic studies based on social media discourse have the aim to mine the sentiments of the population that is experiencing a pandemic, by understanding people's feelings toward the topics related to the disease. For example, Mollema and colleagues found that during the measles outbreak in the Netherlands in 2013 many Twitter users were extremely frustrated because of the increasing number of citizens that refused to vaccinate for, among others, religious reasons. The measles outbreak in the Netherlands began among Orthodox Protestants who often refuse vaccination for religious reasons.

The main distinction among sentiments observed within a given text is between positive and negative feelings. This dimension is commonly defined as emotional valence in cognitive science and cognitive psychology, and more typically defined as sentiment polarity in the machine-learning subfield called sentiment analysis. The exploration of the emotional valence encoded in the tweets has been used in some cases to predict future behavior, for example to predict whether a customer was likely to use a given service a second time, under the assumption that a positive feedback left on Twitter would imply that a client might be more inclined to use that service again. In the case of political messaging during electoral campaigns, positive feedback might correlate with voters' support for a specific candidate. In some cases, as pointed out by recent studies, social media analyses during crisis situations may be used to investigate real-time public opinion and thus help authorities to gain insight for quickly deciding for the best assistance policies to be taken (Mathur et al., 2020).

Temporal analyses of sentiments expressed in Twitter data have been previously done on a variety of topics, including the FIFA Confederations Cup in Brazil (Alves et al., 2014), the changes in voters' feelings during the US elections (Paul et al., 2017) and the changes of sentiments on a monthly, daily and hourly level across different geographical regions (Hu et al., 2019).

## Opinions

As suggested by Liu (2010), facts are objective expressions about events, entities, and their properties, whereas opinions are usually subjective expressions that describe sentiments, appraisals, feelings toward events, entities, and their properties. Research on subjectivity detection, that is, the distinction between texts that express opinions and texts that express facts is becoming increasingly central in various fields, such as computer science, journalism, sociology, and political science (see Chatterjee et al., 2018 for a review). The reasons for this interest are varied. There are business-related issues, such as companies interested in understanding whether consumers have strong opinions toward a specific brand or whether instead they are indifferent. Politics is another field where many use data to understand whether a specific candidate triggers opinions or leaves voters indifferent.

Distinguishing fact-based and opinion-based texts in social media is an operation usually performed by different types of analysts to fulfill different goals. Detecting fact-based texts in the wild (thus filtering out opinion-based texts) is an operation that can be performed by analysts interested in detecting events and capturing factual data, for the automated and fast identification of (for example) breaking news from social media streams. Conversely, detecting opinion-based texts in the wild is an operation that enables analysts to capture users' beliefs and feelings. This is usually done by companies to develop marketing strategies toward their brand. In both types of tasks, Twitter has been used as a valuable resource of linguistic data for fact and opinion data mining (Li et al., 2012).

Subjectivity detection is a major subtask involved in sentiment analysis (Chaturvedi et al., 2018). Before analyzing the positive and negative feelings involved in a corpus of texts, those texts that have a neutral connotation, that is, those texts that are not subjective, need to be filtered out (Liu, 2010). This is usually done in order to ensure that only opinionated information is processed by a classifier that can distinguish between positive and negative feelings. A thorough review of the methods and the challenges involved in distinguishing between facts and opinions for sentiment analysis lies beyond the scope of the present paper (but see Chaturvedi et al., 2018 for a literature review). The following heuristic might summarize how subjectivity and sentiment polarity are related to one another: the more a text includes words that are loaded with (positive or negative) emotional content, the more that text is arguably subjective, as it expresses personal opinions, beliefs, and sentiments toward a specific topic. Conversely, texts that feature neutral words, not loaded with emotions, are likely to be more informative and objective.

## Framing

In cognitive linguistics and communication sciences, and in particular in metaphor studies, public discourse is often analyzed in relation to different figurative and literal communicative framings (Burgers et al., 2016). A frame is hereby defined as a selection of some aspects of a perceived reality, which taken together make a standpoint from which a topic can be seen. Such a standpoint is "constructed to promote a particular problem definition, causal interpretation, moral evaluation, and/or treatment recommendation for the item described" (Entman, 1993. p. 53). Within this definition of framing, metaphors may be used to establish a perspective on a given topic. In health-related discourse, for example, "war-metaphors" are often used to talk about illnesses and treatments. For instance, in a pioneering work, Sontag and Broun (1977) described and criticized the popular use of war metaphors to talk about cancer, a topic of research recently investigated also by Semino et al. (2017). Their argumentation suggested that the use of military metaphors bears negative implications for clinical patients (see also Hendricks et al., 2018). Nevertheless, military metaphors are widely used and highly conventionalized, for their ability to provide a very effective structural framework that can be used to communicate about abstract topics, usually characterized by a strong negative emotional valence. Military metaphors, as suggested by Flusberg et al. (2017) draw on basic knowledge that everyone has, even though for most people this is not knowledge coming from first-hand experience. These metaphors are very efficient in expressing the urgency associated with a very negative situation, and the necessity for actions to be taken, in order to achieve an outcome quickly. As recently reported by Wicke and Bolognesi (2020) this frame is also frequently used to talk about Covid-19 on Twitter. As the authors show, the WAR frame (and thus war-related metaphors) is much more commonly used than alternative figurative frames that can be found in the discourse about Covid. The authors also show that the most commonly used lexical units related to the WAR framing are "fighting," "fight," "battle," and "combat." This may be attributed to the stage of the pandemic during which the study was conducted (peaks of the first wave, March–April 2020). As the authors suggest, it could be the case that different stages in the development of the pandemic are characterized by different uses of the WAR framing, in relation to Covid. For example, it could be the case that new lexical units within the WAR framing become frequently used, to express aspects of the sociocultural situation that were previously non-existent. These intuitions are tested in the present study, in section How Does the Figurative Framing of WAR Change Over Time?

# WHICH TOPICS ARE DISCUSSED ON TWITTER IN RELATION TO COVID-19 AND HOW DO THEY CHANGE OVER TIME WITH THE DEVELOPMENT OF THE PANDEMIC?

## Methods
### Data Acquisition
Twitter counts around 152 million active users worldwide (Statista, 2020). Through the publicly accessible Application Programming Interface (API) services, the platform allows analysts to mine the tweets that users post online, in compliance with the privacy regulations set by the platform programmers. According to the official Twitter redistribution policy[1] it is not allowed to share tweets and the metadata associated with them (user's name, date, etc.), but only tweet IDs, user IDs, and other meta-information alone.

Based on the extensive resource of tweet IDs collected by Lamsal (2020), we created a subcorpus of Covid-related tweets. The original dataset of tweets IDs collected by Lamsal contains 3–4 million tweets per day, in English, retrieved from Twitter based on a list of 90+ keywords that are possibly related to Covid, such as "corona," "coronavirus," or "pandemic."[2] This resource contains tweets and retweets, as well as all tweets produced by any tweeter. For the purpose of constructing a balanced, representative, and computationally manageable corpus of tweets stemming from this extensive archive, we sampled 150,000 tweets per day from Lamsal's resource. From each sample we retained only one tweet per user and dropped retweets. Keeping only one tweet per user allowed us to balance compulsive tweeters and less involved Twitter users, thus preserving the representativeness and balance of the language used on Twitter to talk about Covid. The resulting corpus, on which the current analyses were performed, contains 1,698,254 tweets from individual users (without retweets), produced between 20.03.2020 and 01.07.2020.

### Topic Modeling
The topic modeling analysis hereby implemented builds on an approach presented by Wicke and Bolognesi (2020), that uses Latent Dirichlet Allocation (LDA) (Blei et al., 2003). The standard LDA algorithm is an unsupervised machine learning algorithm that aims to describe samples of data in terms of heterogeneous categories. Since the LDA algorithm is unsupervised, the analysts need to specify the amount of topics to be modeled. For example, by specifying $N = 4$, each tweet in the corpus will receive a likelihood to belong to one of four categories automatically identified by the algorithm. The categories are defined by words that have the strongest co-occurrence with each other.

The processing and modeling pipeline can be summarized as follows:

- Stopword removal: the most common English words, e.g. "a," "the," "but," are filtered out, based on established stopword lists (Stone et al., 2011; Wei et al., 2015; NLTK[3]).
- Tokenization: by means of the NLTK Tweet Tokenizer[4]
- Gibbs Sampling: by means of the Mallet library (Rehurek and Sojka, 2010) for Gensim to apply Gibbs Sampling in our LDA training.
- Number of topics: we explored all topic modeling solutions from topic number $N = 2, 3, 4 \ldots$ to $N = 32$, and then a highly granular solution, $N = 64$. Based on the coherence measure of the cluster solutions $C_v$ (Syed and Spruit, 2017) we retained the best solutions.

Internal topic coherence is evaluated through the elbow method: all topic numbers are plotted in relation to their internal coherence and the selected solutions are those in which the function shows a clear bend, suggesting that for the next solution the coherence slope drops significantly. For the purpose of this study, we aimed at picking 4 different cluster solutions that vary in their degree of granularity. We partitioned the data into a smaller and into a larger number of topics in order to see potential differences emerge between a broad analysis and a more fine grained analysis of the topics within our corpus.

### Temporal Analysis
**Figure 1** displays the steps involved in the topic modeling based on the corpus of Covid-related tweets. The 100 groups of daily tweets were fed into the topic modeling algorithm, which provides the probability distribution for each tweet to belong to a certain topic. The result of the temporal modeling analysis is a series of clusters for each topic, for each day in the corpus. Based on these temporal distributions we provide an analysis of the observed patterns co-occurring with events in the news.

## Results
Following the process pipeline depicted in **Figure 1**, we created 32 LDA models, each with a different number of topics. The evaluation of the $C_v$ coherence measure revealed an elbow of the function for $N = 20$ (see a plot of the curve on the Open Science Framework OSF platform repository[5] for this paper). In addition to this, we selected a model that allowed for a broader analysis of the topics, hence a smaller number of (more inclusive) topics. Based on our previous experience with topic modeling and the coherence value function we selected $N = 12$, together with $N = 32$ for the fine-grained solution, and $N = 64$, our most fine grained solution. It should be noted that the LDA algorithm itself involves some degree of randomness and therefore it is likely to obtain different models, even when trained on the same data. Yet, our selection is based on the evaluation of the coherence measure in order to mitigate the statistical randomness.

---

[1]https://developer.twitter.com/en/developer-terms/policy
[2]https://ieee-dataport.org/open-access/coronavirus-covid-19-tweets-dataset

[3]https://www.nltk.org/book/ch02.html#code-unusual
[4]https://www.nltk.org/_modules/nltk/tokenize/casual.html#TweetTokenizer
[5]https://osf.io/v523j/?view_only=63f03e24d48c4d58af1793e0f04ce28b

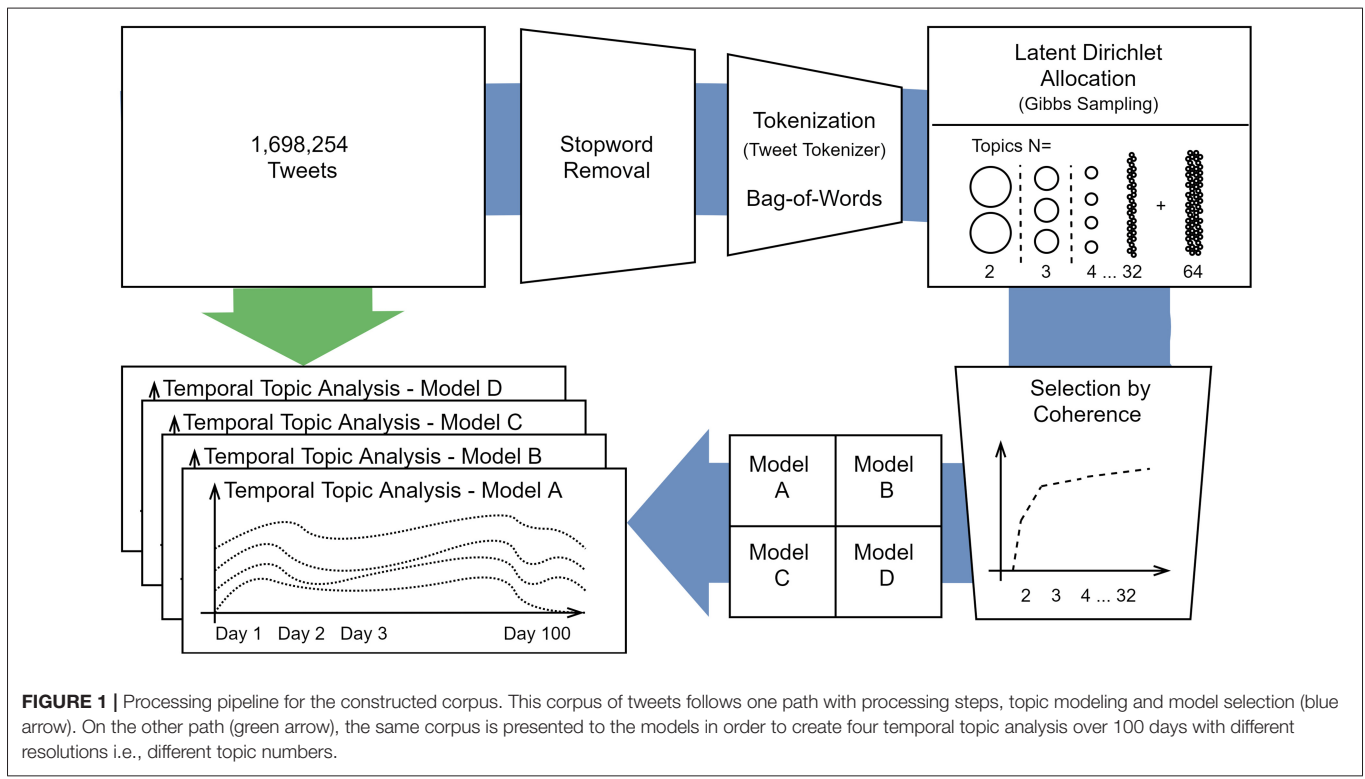

**FIGURE 1** | Processing pipeline for the constructed corpus. This corpus of tweets follows one path with processing steps, topic modeling and model selection (blue arrow). On the other path (green arrow), the same corpus is presented to the models in order to create four temporal topic analysis over 100 days with different resolutions i.e., different topic numbers.

Model A ($N = 12$), B ($N = 20$), C ($N = 32$), and D ($N = 64$) are stored in the OSF online repository, where the plots can be dynamically explored using an interactive web-service that we created using pyLDAvis[6] (Sievert and Shirley, 2014).

In order to capture the temporal dynamics involved in the discourse, the groups of tweets collected for each of the 100 days (an average of 12,598 tweets per day, once the corpus is filtered for retweets and unique users) were fed into the topic modeling algorithm. **Figures 2–5** illustrate the four analyses displaying the temporal line on the horizontal axis, the topics as different chromatic shades on the vertical axes, and the proportion of tweets within each topic (day by day) is represented by the colored areas. The labels in the legend for each topic consist of the top 3 or 4 most important words within each topic. These are visible in larger fonts in the interactive versions of these topic modeling solutions.

**Figure 2** displayed the less granular topic modeling analysis ($N = 12$), which is likely to capture broader and more generic topics associated with the discourse about Covid-19. Three main observations can be made, based on the changes in the colored areas. In **Figure 6** we have highlighted those three bands, which occurred roughly in the first week of April, fourth week of April and fourth week of May. The pandemic-related events we correlate with the results of the topic modeling are being informed by official statements released by the World Health Organization (2020).

In the first week of April, we observe a change in the topics "stop|government|money" and "year|fucking|years|months." We interpret this in relation to a major event that took place on the second of April: A record of 6.6 million Americans filed claims for unemployment.[7] As a consequence, we argue, the first week of April had a strong impact on people's opinion on continued/stopping government financial aid throughout the upcoming months/year.

The fourth week of April shows a strong increase for the topic "mask|masks|vaccine|face". The following additional keywords are associated with this topic: "kill," "cure," "disease," "human," "wear," "person," "treatment," "wearing," "body," "science," "light," "research," "sense," "common," and "study." Comparing this topic with the news reported by the press, it appears that on 23rd April Donald Trump suggested (ironically or otherwise) that coronavirus might be treated by injecting disinfectant or by UV lights.[8] It is likely that this comment triggered the increased discussion on Twitter about common sense, science, and effective treatment (such as masks, vaccines, face masks).

The 20 topics solution shows the following trends:

- The topic marked as "home|stay|lockdown" displays a large portion of tweets in March; then the concentration decreases, to finally increase again in June. These trends might be related

[7]https://www.bls.gov/opub/ted/2020/unemployment-rate-rises-to-record-high-14-point-7-percent-in-april-2020.htm?view_full
[8]https://www.bbc.com/news/world-us-canada-52407177

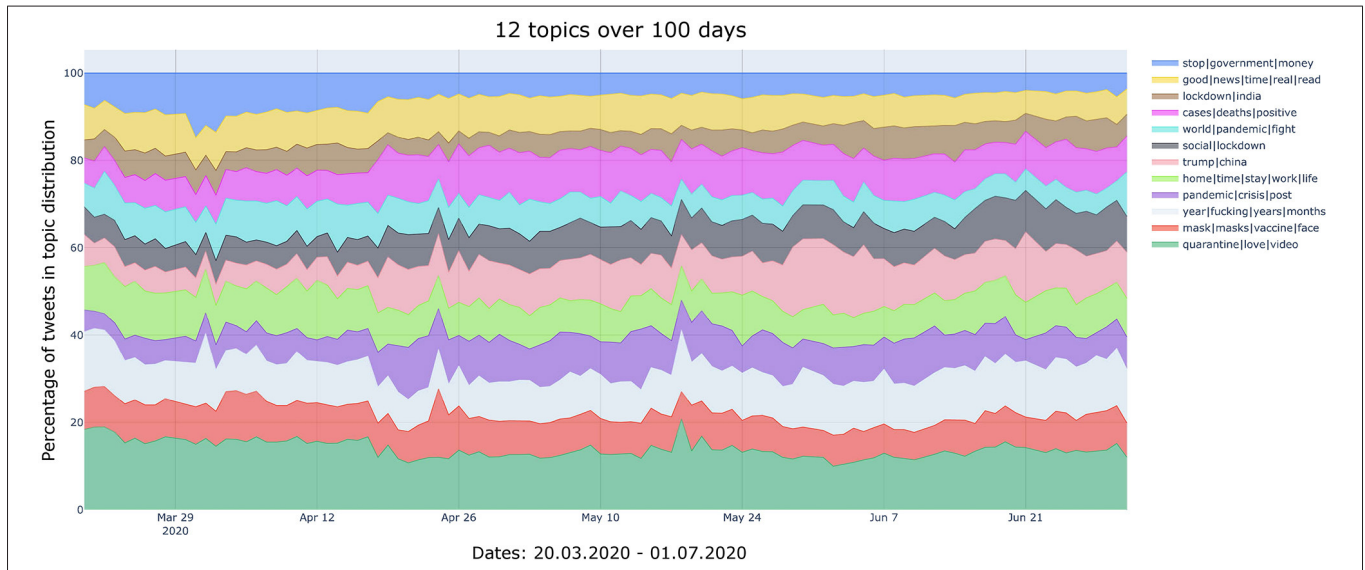

**FIGURE 2** | Temporal development of the topics (*N* = 12 topics, 100 days). Interactive version: https://bit.ly/3cfDq0V.

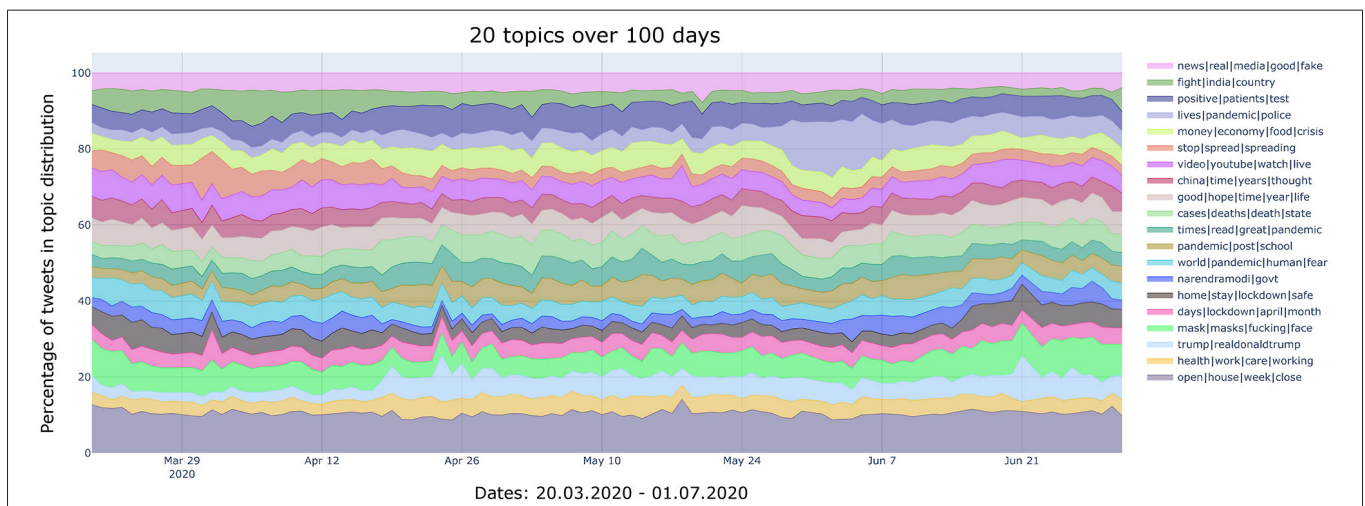

**FIGURE 3** | Temporal development of the topics (*N* = 20 topics, 100 days). Interactive version: https://bit.ly/2FNACwb.

to the "Stay at home" guidelines issued by the WHO on 12th March and updated and extended on 29th March.

- In the first half of April there is a concentration of tweets in the topic labeled as "stop|spread|spreading." We interpret this as a reaction to some notifications issued by the WHO, such as the confirmation of over 1 million cases of COVID-19 reported on 4th April 2020, the updated guidance on how to use masks on 6th April, and the publication of a draft landscape of COVID-19 candidate vaccines on 11th April. In relation to this, the increased concentration of tweets in the topic labeled as "positive|patients|test" around the end of April/beginning of May, might be related to the fact that in this period the USA became the first country in the world to hit 1M cases.

- A substantial concentration of tweets is observed in early April, in the topic labeled as "fight|India|country." This is when the virus started to spread exponentially in this country.

- An increase of tweets in the topic labeled with the keyword "narendramodi" (Narenda Modi is the Indian Prime minister) can be observed around the first half of June, when the spreading of the virus was particularly fast in this country, and the Prime Minister was appearing often in the media, with messages related to the pandemic. Moreover, in the beginning of June he held summits with authorities in France and in the USA. Finally, another peak can be observed around the 28th June. We interpret this as an anticipation of the major event that took place on the 30th June, when Modi addressed the whole nation with a strong message, explaining that people had become more irresponsible and careless about COVID-19 prevention guidelines since the start of their first "Unlock 1.0."[9]

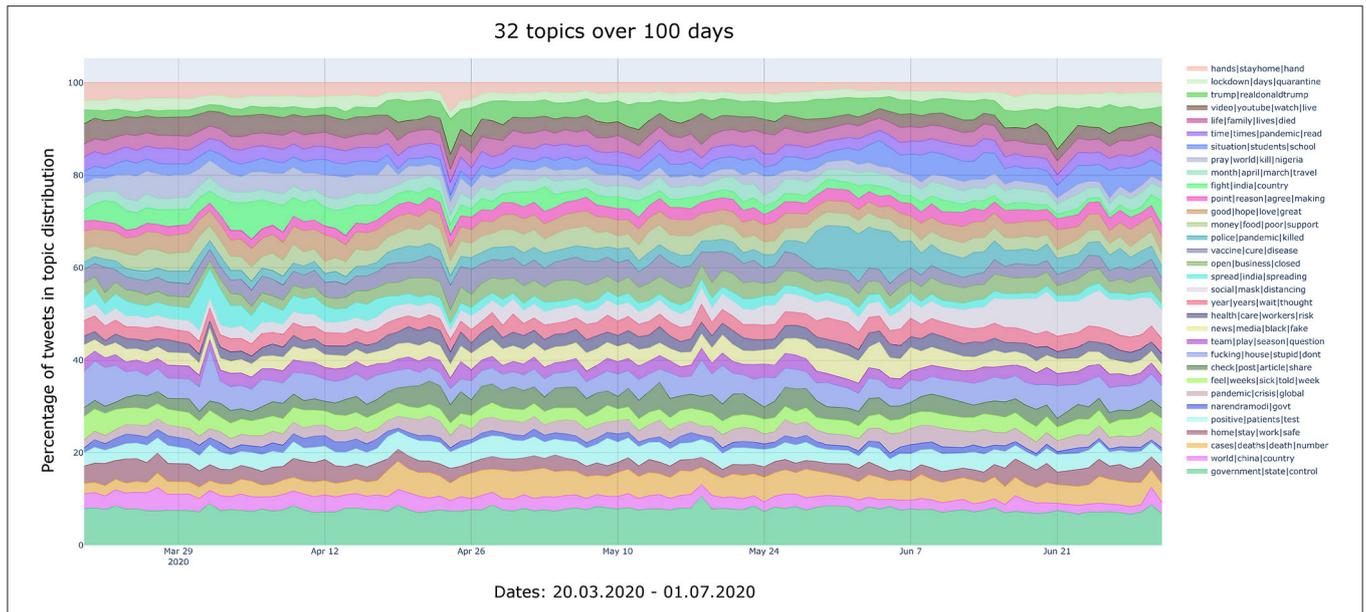

**FIGURE 4 |** Temporal development of the topics ($N = 32$ topics, 100 days). Interactive version: https://bit.ly/35RotkG.

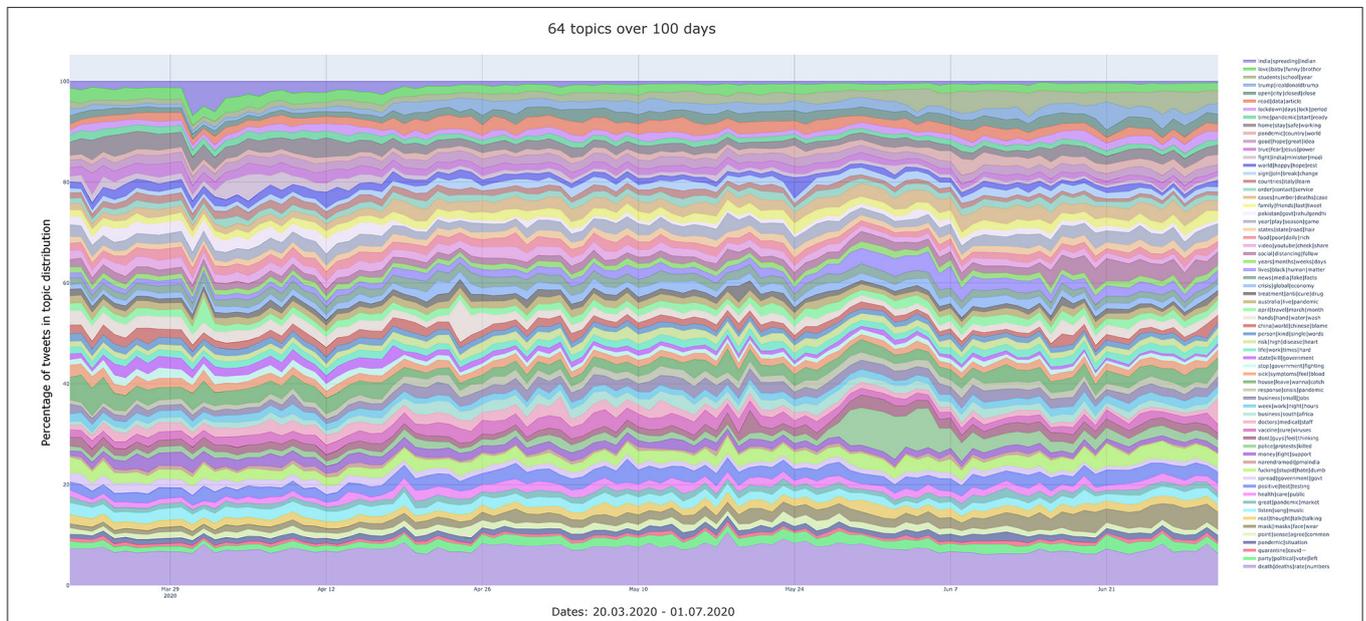

**FIGURE 5 |** Temporal development of the topics ($N = 64$ topics, 100 days). Interactive version: https://bit.ly/3kx6XGo.

- The topic labeled as "Trump" displays three main peaks of tweets, on 24th and 26th April, as well as on 21st June. The first two dates correspond to the days that followed the statement by Donald Trump in which he was floating the idea (ironically or otherwise) of ingesting disinfectants as a potential coronavirus treatment. The latter date corresponds to the date in which he held his first campaign rally since the US coronavirus lockdown began, in front of a smaller than expected crowd in Tulsa, Oklahoma.

[9]https://www.oneindia.com/india/key-takeaways-from-pm-modi-s-address-to-nation-on-june-30-3112740.html

- Finally, a crucial topic, previously undocumented in the $N = 12$ solution, becomes particularly relevant around the end of May. This topic is labeled as "lives|pandemic|police" and

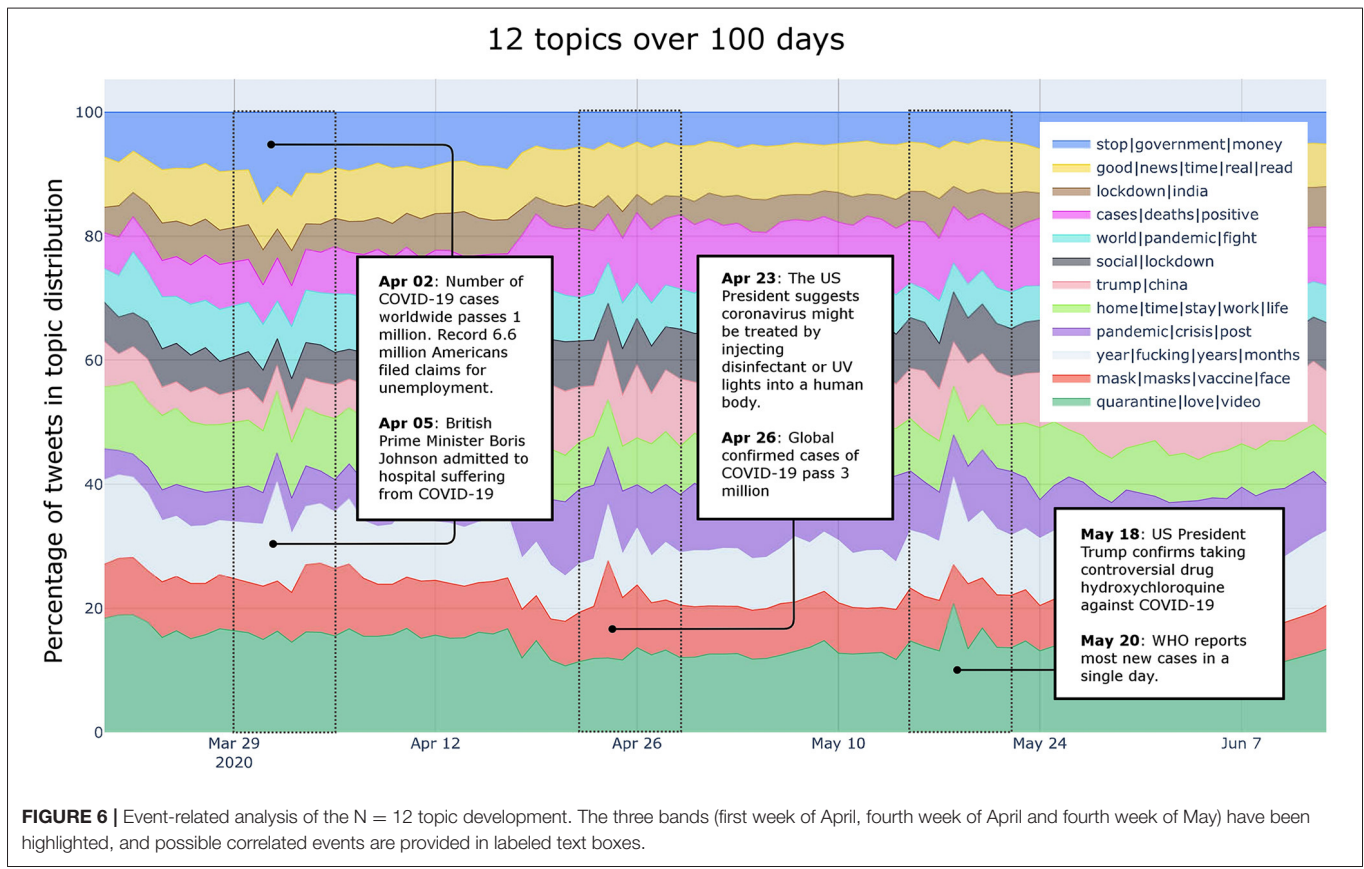

**FIGURE 6** | Event-related analysis of the N = 12 topic development. The three bands (first week of April, fourth week of April and fourth week of May) have been highlighted, and possible correlated events are provided in labeled text boxes.

its appearance collides with the murder of African American George Floyd, by police officers. This is further described in relation to the most fine-grained topic solution, $N = 64$.

We also acknowledge a couple of concentrations of tweets on 18th May in the topics labeled, respectively, as "video|Youtube|watch|live" and "open|house|week|close," for which we were not able to identify any specific event that may be associated with this specific date.

The $N = 32$ cluster solution, which is more fine-grained than the previous solutions described above, displays a few interesting trends in addition to those emerging from previous analyses:

- A peak of tweets can be noted on 18th May on the topic "government|state|control," which may precede by a couple of days the official CDC announcement (probably leaked by the press a few days before its official release). This peak coincides with the introduction of a Community Mitigation Framework that includes updated guidance for communities, schools, workplaces, and events to mitigate the spread of COVID-19.[10] This is also related to a peak in the topic "vaccine|cure|disease," observed on 19th May.
- A peak of tweets on the topic "fucking|house|stupid" is observed on 1st April, when the WHO issued a report with

specific guidelines for Public Health and Social Measures for the COVID-19 Pandemic.
- A substantial increase of tweets within the topic "social|mask|distancing" observed in June, concurrently with the gradual reopening of various countries and the need to remind people to keep safe distance.

Finally, the $N = 64$ topic development provides the most fine-grained analysis of the topics. These are reported in **Figure 7**. Here it can be observed that many events already mentioned in the previous analyses are captured also by this model. Moreover, the more detailed analysis reveals an increase of the topic "india|spreading|indian" around the 31st March, in addition to the peaks observed and described in the previous models. Around this time India and Pakistan intensified their efforts to contact-tracing participants of the Tablighi Jamaat coronavirus hotspot in Delhi with more than 4,000 confirmed cases.[11] We therefore take this feature of the $N = 64$ model as a good example of the topic modeling capturing local events with a greater number of topics.

Focusing on the latter half of the 100 days, we can explore how apparently unrelated topics are entering the Covid-19 discourse. For example, on Saturday 23rd May we can observe an increase for the topic labeled "world|happy|hope|rest." Upon closer investigation of the topic model, we identify the related topic words: "allah," "pray," "save," "bless," "month," and "protect". This might refer to the end of Ramadan and the Eid Al-Fitr (festival of breaking the fast), since it correlates directly with the words "month," "pray," "allah," and "bless." These topics at first sight might appear to be unrelated to COVID-19. However, these are extracted from tweets that feature one or more of the COVID-19-related keywords. Tweeters are therefore likely to have expressed some connection between these events and the pandemic.

---

[10]https://www.cdc.gov/coronavirus/2019- ncov/downloads/php/CDC- Activities-Initiatives-for-COVID-19-Response.pdf

[11]https://www.washingtonpost.com/world/asia_pacific/india-coronavirus-tablighi-jamaat-delhi/2020/04/02/abdc5af0-7386-11ea-ad9b-254ec99993bc_story.html

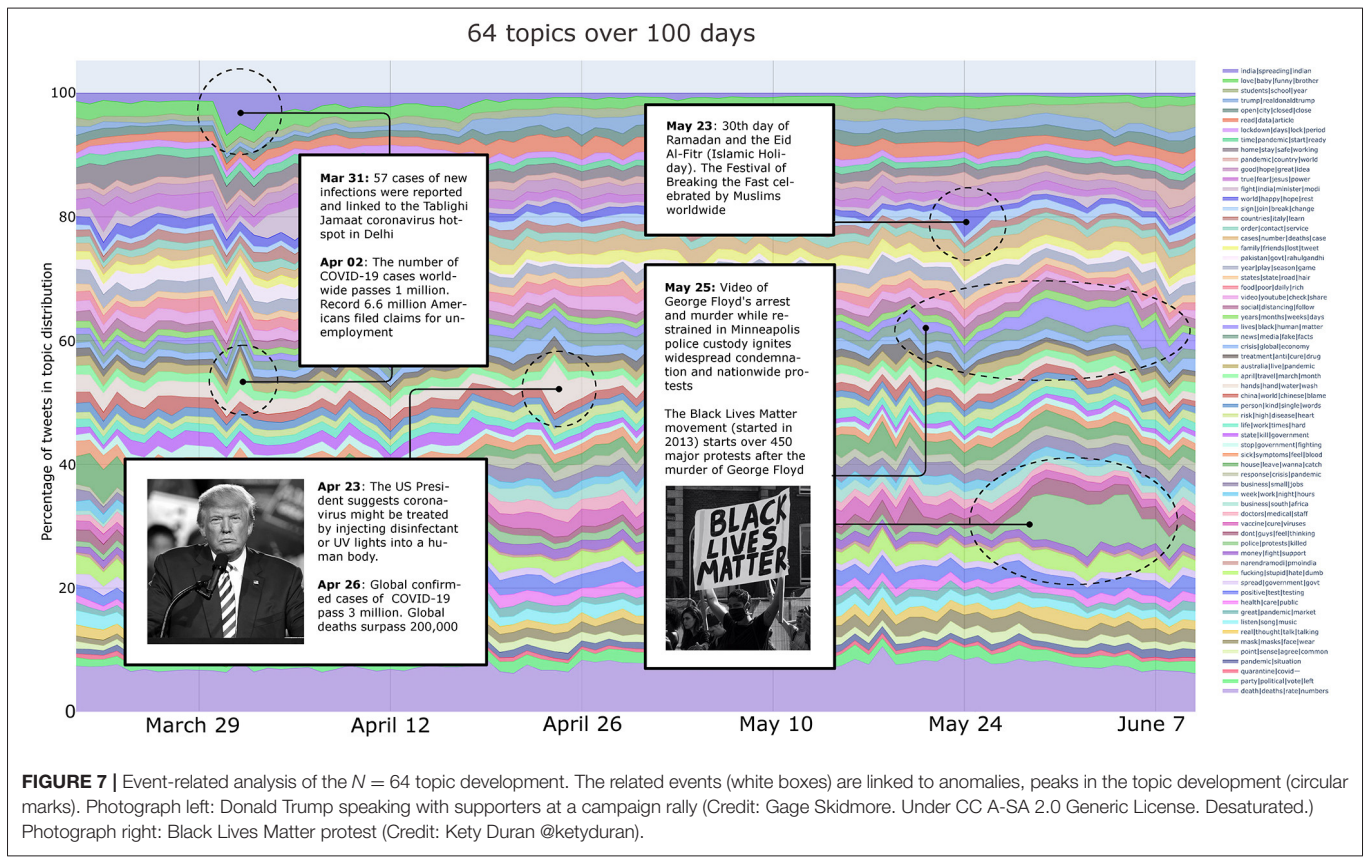

**FIGURE 7 |** Event-related analysis of the $N = 64$ topic development. The related events (white boxes) are linked to anomalies, peaks in the topic development (circular marks). Photograph left: Donald Trump speaking with supporters at a campaign rally (Credit: Gage Skidmore. Under CC A-SA 2.0 Generic License. Desaturated.) Photograph right: Black Lives Matter protest (Credit: Kety Duran (@ketyduran).

Finally, on the 25th May a video of African American George Floyd's arrest and murder while under restraint in Minneapolis police custody shows the moment when he was pinned to the ground by a police officer for 8 min and 46 s. This video ignited widespread condemnation and Nationwide protests in the U.S. In **Figure 7**, despite the large number of topics displayed, we can clearly observe how this event has affected the discourse about the COVID-19 pandemic. On the 25th May the topic labeled "police|protests|killed" shows a great concentration of tweets. The most important words in this topic are: "police," "protests," "killed," "dead," "protest," "thousands," "killing," "mass," "riots," and "protesting." Although it might appear that this topic is related to a different set of events, it is worth remembering that all the tweets on which the analysis is performed showcase a keyword associated to Covid-19 and its variants. At the same time, the graph shows an increase for the topic labeled "lives|black|human|matter," which indicates how the Black Lives

Matter (BLM) movement has gained momentum after the murder of George Floyd. Also, this topic is discussed in relation to Covid-19.

## Discussion

The topic analyses show different trends. The less granular analysis, based on a limited number of topics ($N = 12$) shows a macro distinction into topics of discussion, where general themes emerge. Conversely, the more the number of topics increases, the more the tweets are partitioned into smaller clusters, which are more thematically coherent and seem to capture more specific events reported by the media and discussed by the Twitter users. Overlaps can be observed as well, across the various topic modeling solutions, with some trends emerging in generic as well as in more granular topic models. Nevertheless, we showed that a multiple approach to the data partitioning provides a better view into the data trends.

The explorative nature of the topic modeling approach allows analysts to mine large collections of linguistic data in a bottom-up manner, to observe tendencies of language use emerging from authentic texts. However, it should be acknowledged that the association of linguistic trends to specific events reported in the news, is an interpretative process subject to a degree of variability. Although we based our interpretations on the keywords emerged from the topic models and on major sources of information such as the WHO and the CDC websites, in principle it could be argued that different (but related) events reported in the media might have explained the changes in topics observed though our topic modeling analyses.

## WHAT VALENCE EMERGES FROM THE TWEETS ABOUT COVID-19 AND HOW DOES IT CHANGE OVER TIME?

### Methods

For each of the 100 days in our corpus, the average polarity of the words used in the tweets was assessed using the TextBlob library (Loria et al., 2014). The obtained polarity score was a numeric value within the range $[-1.0, 1.0]$ where $-1.0$ represents a very negative sentiment and 1.0 represents a very positive sentiment.

TextBlob's sentiment analysis (previously applied to Twitter data, see Hawkins et al., 2016, Reynard and Shirgaokar, 2019) is based on the Pattern library, which uses a lexicon of hand-tagged adjectives, with values for polarity and subjectivity (De Smedt and Daelemans, 2012).

After calculating the sentiment scores, we identified the most appropriate function to describe the change of sentiment emerging from the distribution of the tweets over time. For that, we started by using a polynomial regression ($f(x) = \beta_0 + \beta_1 x^1 + \beta_2 x^2 + \ldots + \beta_N x^N + \varepsilon$ where $\varepsilon$ is an unobserved random error). Specifically, we performed an ordinary least-squares regression for an increasing polynomial degree until we explained much of the data variance with significant confidence.

### Results

**Figure 8** reports the average polarity scores for each of the 100 days in the corpus. A linear function ($f(x) = 0.0622 - 0.0001x$)

could only explain 8.8% of the variance of the data ($R^2$: 0.088, $p > 0.003$, F-statistic = 9.480). Therefore, we modeled a polynomial function of second degree (**Figure 8**, curved dashed black line) with $f(x) = 0.0513 + 0.0006x - 0.000007x^2$. Assuming this non-linear correlation between time and polarity is a better fit ($R^2$: 0.356, $p < 0.001$, F-statistic = 26.81), and explains 35.6% of the variance. Notably, higher polynomials provide even better fit, yet they do not serve our investigation to identify a simple trend and can overfit our data.

As **Figure 8** shows, the sentiment scores are displayed on an inverted U shape, which suggests that the sentiments expressed by the tweets are increasingly positive from the beginning of our timeframe to averagely the middle of the interval (therefore, from 20 March to the beginning of May) and then drop toward the negative end of the valence spectrum.

### Discussion

The overall sentiment polarity over 100 days emerging from the tweets in the corpus is slightly positive ($>0$). The polynomial regression indicates that the average sentiment is increasingly more positive during the first 40 days of the pandemic, while it drops dramatically in the second half of the timeframe. We interpret this trend in the following way. Within the first month of the pandemic, the general attitude of the population tended to be slightly optimistic. These first 40 days refer to the last 10 days of March and the whole month of April. Many countries during this period were in lockdown, and despite the fear toward the unknown situation, a positive attitude, even though only verbally expressed in the tweets, might have been a form of mutual encouragement. This is the period in time

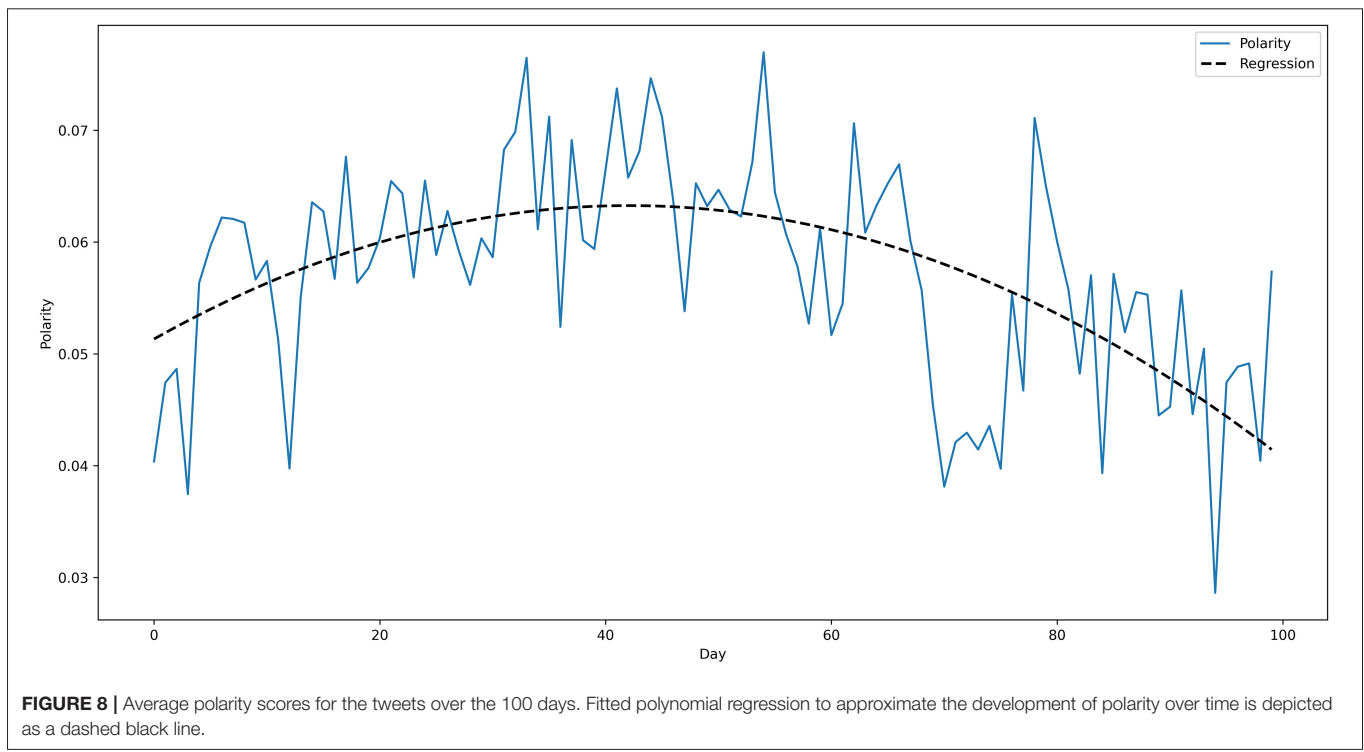

**FIGURE 8** | Average polarity scores for the tweets over the 100 days. Fitted polynomial regression to approximate the development of polarity over time is depicted as a dashed black line.

where a series of positively framed hashtags began to emerge, such as #StayHome, #FlattenThe Curve, #StayHomeSavesLives, encouraging people to embrace the difficult situation and hold on tight, in order to fight together against the virus. This is the type of attitude expressed by the collective mind, emerging from Twitter. An example that summarizes this attitude is the following tweet, which was anonymized, in compliance with the Twitter regulations:

*"Probably for the best, don't need everyone's grubby hands digging in them spreading corona virus around. #brightside"* (20th March 2020)

It should be noticed that during this first phase (March/early April), the financial and societal repercussions of the lockdown were not yet as obvious and impactful as in the latter half of the 100 days (May/June). During the later dates in fact, the attitude in the collective mind of the population dropped substantially, toward a much more negative end. The following tweet exemplifies this trend (polarized toward the negative end):

*"Bullshit!!! Our country is the worst in the world for the pandemic because of you, not China. Stop blaming everyone else & trying to defect blame. The state of our country is your fault & yours alone. RESIGN!!!"* (1st July 2020)

This trend toward negative sentiments arguably signals a change in the general well-being of the population in relation to the current pandemic. In line with previous research, we believe that keeping track of the polarity of the sentiments expressed on social media may be beneficial to health practitioners and politicians. They could potentially understand what will be the most effective measures to contrast the degeneration of the general well-being of the population. For example, it may be suggested that restrictive measures such as hard lockdowns imposed in during period when the general feelings are particularly negative, may lead to extreme and undesirable individual and collective actions.

As a caveat of our analysis, it should be acknowledged the fact that TextBlob (as many other lexical approaches to sentiment analysis) does not distinguish genuine positive sentiments from sarcastic ones. This is an open issue in sentiment analysis, and a major bottleneck in machine learning in general, which is currently being tackled by scholars that are developing tools for the automatic detection of sarcasm and irony (Ghosh and Veale, 2016, Reyes et al., 2013). The issue is exemplified by the following tweet:

*"Best healthcare system in the world 😊 😕"* (1st July 2020)

While the average polarity of this tweet leans toward the positive end, the pragmatics of this message and the use of specific emoji that express negative emotions such as skepticism and frustration, suggests that the user is being sarcastic. Therefore, the tweet would need to be interpreted as emotionally negative. The real polarity of the tweet is, in this case, carried by the emoji, rather than by the verbal text. Further research might need to disentangle genuinely positive tweets related to the pandemic from tweets that express sarcastic comments toward the current situation, which might need to be classified as negative even though they feature words loaded with positive feelings.

## HOW DOES THE SUBJECTIVITY OF THE TWEETS (I.E., OPINION-BASED FOCUS, VS. THE OBJECTIVE FACT-BASED FOCUS) CHANGE OVER TIME?

### Methods
The TextBlob tool for Sentiment Analysis provides a subjectivity score in addition to the polarity score. The subjectivity score is in the range [0.0, 1.0] where 0.0 is very objective (facts) and 1.0 is very subjective (opinions). As for the previous analysis, we averaged the subjectivity scores for each day and then we identified the most appropriate function to describe the changes in subjectivity. For that, we used an ordinary least-squares regression for an increasing polynomial degree until the function could explain a large portion of data variance with significant confidence.

### Results
The result for the subjectivity score over 100 days is depicted in **Figure 9**. A linear regression for the subjectivity scores indicated a good fit. The linear regression $f(x) = 0.3452 + 0.0003x$ showed $R^2 = 0.69$, F-statistics $= 1.06e{-}26$ and highly significant $p < 0.0001$, implying that about 69% of the variability in subjectivity scores is explained by passing days.

Here we report an exemplar tweet produced on 20th March, which is associated with a value of 0 subjectivity, which means that the text is likely expressing facts rather than opinions, and it is objective rather than subjective:

*"things corona has done to us: made us wash our hands canceled allergy season/ coughing and sneezing otherwize people think we are possessed"* (20th March 2020)

This tweet expresses substantially factual information, even though colored with a quite humoristic nuance. This type of objectivity can be compared with the following tweet, also produced on 20th March, and also colored with a humoristic nuance, in which the author indicates a different type of consequence associated with Covid-19:

*"a guy just messaged me and said "if corona doesn't take you out can i?" literally the worst thing coronavirus has done is fuel men's terrible flirting rhetoric"* (20th March 2020)

The latter tweet is one of the few examples of tweets associated with high subjectivity scores, produced early in our time frame (20th March). Comparing the former with the latter tweet, it becomes clear why the former tweet scores low on subjectivity: while in the first case Covid-19 is associated with washing hands and avoiding sneezing (even though canceling allergy season is a hyperbolic statement), in the second case Covid-19 is associated with fueling men's terrible flirting rhetoric. The negative judgment toward specific flirting techniques is arguably subjective, and so it is the attribution of its responsibility to Covid-19.

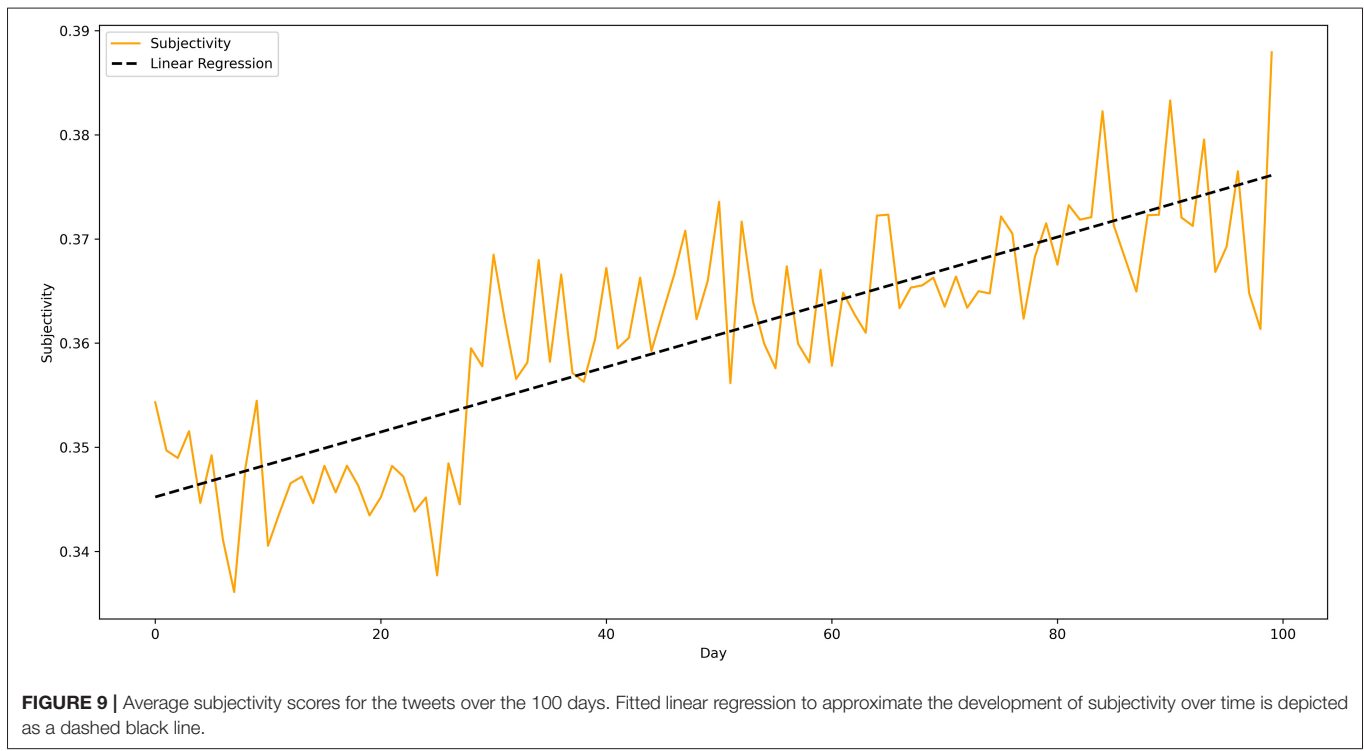

**FIGURE 9** | Average subjectivity scores for the tweets over the 100 days. Fitted linear regression to approximate the development of subjectivity over time is depicted as a dashed black line.

An example of tweet that scores in the middle of the subjectivity scale (subjectivity = 0.5), posted on 10th May, in the middle of our time frame, is the following:

*"I want to say Trump but Covid-19 is a bigger threat to the world"* (10th May 2020)

This tweet, as most tweets, uses elliptical language in which words and punctuation are omitted. From a pragmatic perspective this opens up the field to various possible interpretations. In our opinion, this tweet seems to express two meanings: [1]. Trump is a big threat to the world, and [2]. Covid-19 is a bigger threat to the world than Trump. These two meanings are expressed linguistically in a way that helps the tweeter build a rather sophisticated argumentation. First, the tweeter suggests to the reader that she thinks that Trump is a big threat to the world. She expresses this statement as an opinion ("I want to say..."). Then, the tweeter introduces the second part as an assertive statement (Covid-19 is a bigger threat to the world than Trump). Here the tweeter wants the reader to take this statement as an objective, factual piece of information. In this construction, the first statement contributes to build the perceived objectivity expressed in the second statement. As a result, the average subjectivity of the tweet scores a medium value.

Finally, an example of a tweet produced on 1st July (end of the time frame), associated with a high subjectivity value (subjectivity = 1) is the following:

*"maybe being stupid is a pre-existing condition that makes you susceptible to the corona virus?"* (1st July 2020)

In this tweet, the high subjectivity is given by the fact that the tweeter introduces the statement with a "maybe," which signals a possibility, which in turn suggests that this is an opinion, and then proposes stupidity (a highly subjective human trait) as a pre-existing condition possibly associated with Covid-19.

## Discussion

The subjectivity analysis displayed in **Figure 9** shows that with the development of the pandemic, Twitter users tend to focus more on their own introspections and to express increasingly more (subjective) opinions than (objective) facts. In other words, the subjectivity of the tweets increases linearly, as a function of time, as we had hypothesized.

Taken together with the previous analysis, the subjectivity and polarity trends suggest that in the beginning of the pandemic Twitter users tended to communicate and rely on facts. They expressed little emotional content, which was initially averagely negative. It is possible that this was characterized by feelings of fear toward the unknown situation. This initial negativity was followed by a positive trend concurrent with the lockdown measures that characterized several English-speaking countries. During these weeks, despite the difficulties experienced by many families, it is possible that a sense of community and need for mutual encouragement led Twitter users to post tweets that were on average more positively valenced than those posted in the previous weeks. The trend dropped again toward the negative end of the scale around the end of April, through May and June.

This phenomenon is concurrent with the gradual reopening of many States (which spans from mid-April to mid-June). During these warmer months, the pandemic developed in many English-speaking countries (notably in the USA), and the number of active cases increased exponentially, showing that the lockdown measures adopted provided only a temporary relief. This negative slope in the polarity of the tweets is concurrent with a substantial increase in subjectivity, and therefore in the increasingly large number of tweets that express opinions, personal beliefs, and introspections, rather than facts.[12]

## HOW DOES THE FIGURATIVE FRAMING OF WAR CHANGE OVER TIME?

### Methods

To understand whether the use of the WAR frame changes over time we used the list of war-related lexical units described in Wicke and Bolognesi (2020), which was created using the web-service relatedwords.org and the MetaNet repository entry for "war."[13] We then computed the distribution of war related terms in our corpus. Since we had a varying number of tweets per day, we randomly sampled 9,500 tweets for each of the 100 days. For each sample of 9,500 tweets, we counted the occurrences of war terms. Finally, we looked at how the occurrence specific war terms changed over time.

Through a regression analysis we described the distribution of war-related terms over time. Subsequently, we defined three time intervals that allowed us to investigate different events and their effect on some selected war-related terms. Three intervals were identified, of 30, 30, and 40 days each, respectively. These intervals overlap with the intervals that characterized two peaking waves of infections observed on the WHO Coronavirus Disease Dashboard (covid19.who.int). **Table 1** describes the intervals and related dates. The methodology hereby adopted has been adapted from a study conducted by Damstra and Vliegenthart (2018).[14]

For selection of terms, we created two constructs. Since the occurrence of war-related terms in the COVID-19 discourse follows a Zipf distribution (Wicke and Bolognesi, 2020), on one hand we considered only the four terms that occur the most often as our first construct, which we labeled as *MostCommon*. This includes the terms "fighting," "fight," "war," and "attack." On the other hand, by looking at the most common words occurring in each of the four time intervals, we noticed that the war-related terms "riot," "violence," "military," and "soldiers" appeared to follow a different temporal distribution than the other terms. Hence, we grouped these four terms in a second construct labeled *EventRelated*. We then analyzed the distributions of these two constructs with a 2-way ANOVA statistical test in

---

[12]In this supplementary plot (https://cutt.ly/ekYrv3l), stored in the OSF online repository of this project, we show that tweets that express highly negative sentiments and highly positive sentiments (right area of the plot), are also associated with high levels of subjectivity (explored in further visualization of the analysis), specifically the polarity and subjectivity.

[13]https://metaphor.icsi.berkeley.edu/pub/en/index.php/Frame:War

[14]We are thankful to Christian Burgers for advising this study, which inspired our analysis hereby reported.

---

**TABLE 1 |** The three chosen intervals for the detailed analysis of correlation between occurrences and war-related terms over 100 days.

| Name | Time interval | Description |
|---|---|---|
| Interval 01 | 20.03–19.04.2020 | The first global rise of infections with a peak of daily confirmed cases (89,335) on the 11th April and a peak of daily deaths (12,430) in the 17th April[a] |
| Interval 02 | 20.04–19.05.2020 | The first global relaxation of infections. 29th April shows the lowest death number on a single day (67,167) since the beginning of the pandemic. |
| Interval 03 | 20.05–01.07.2020 | The second global rise of infections with the highest number of daily new infections since the start of the pandemic (191,028). |

[a]https://covid19.who.int/.

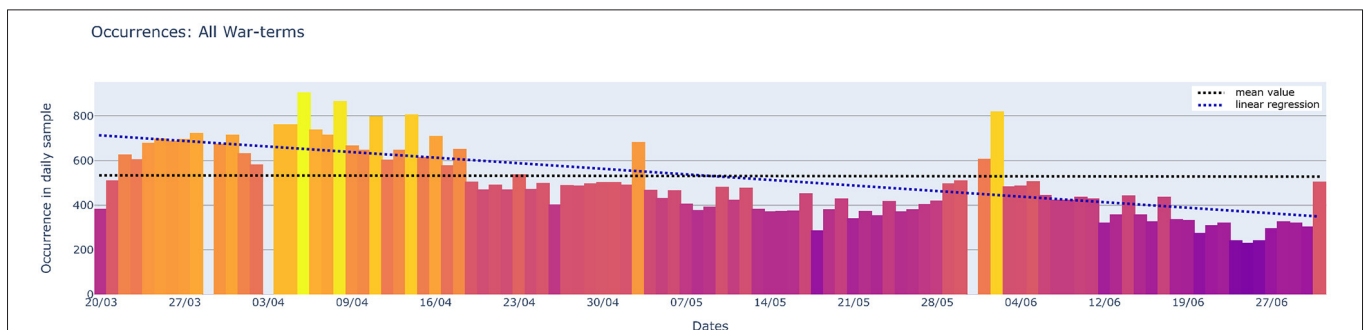

**FIGURE 10 |** Occurrences of war related terms day by day. Brighter color values indicate higher number of occurrences. The three missing bars in the graph correspond to the three missing dates from Lamsal's dataset.

order to test for differences between these two variables (time intervals, constructs). Finally, we performed *post-hoc t*-tests with a Bonferroni correction in order to analyze the effect sizes using Cohen's D, to validate our intuitions about the distribution of war related terms in relation to the timeline.

## Results

The total number of sampled tweets is 950,000 (100 days with 9,500 sample size) and the total occurrence of tweets with war-related terms is 53,214, which corresponds to 5.60% of all sampled tweets. The linear regression $f(x) = 707.26 - 3.5379x$ with $R^2 = 0.513$, F-statistics=103.1 and significant *p*-value, showed that about 51.3% of the variability of occurrences is explained by passing days. The linear regression (blue dotted line in **Figure 10**) can explain the variance of the data better than a linear model of the occurrences with the mean as a constant (horizontal, black dotted line), assuming the occurrences are equal throughout the 100 days (see **Figure 10**).

The results of the 2-way ANOVA showed significant differences across all variables time intervals and constructs. The results are summarized in **Table 2**.

The *post-hoc t*-tests showed significant differences in all tested variations except for one case, as illustrated in **Table 3** (row in bold). The test evaluates the contrast between the constructs with respect to the time intervals. The difference in constructs irrespective of the time intervals is shown in the first row. The difference in the time intervals irrespective of the constructs is shown in the following three rows. Whereas, the contrast for the constructs between the time intervals is shown in the proceeding rows. Except for the difference between time interval 2 and interval 3, irrespective of the construct, all other differences are significant with $p < 0.005$. Notably, Cohen's D is interpreted in direction of column A with 0.2 to be considered a small, 0.5 a medium, and anything >0.8 a large effect size.

The distribution of occurrences over time can be observed for the construct of *MostCommon* terms in **Figure 11**. There, the time intervals are indicated with vertical black lines. The figure shows how the most frequent lexical entries within the WAR frame ("fight," "fighting," "war," and "attack") are used day after day in the tweets about Covid.19. Overall, the frequency of usage appears to decrease over time. The distribution of occurrences over time for the construct of the *EventRelated* terms can be observed in **Figure 12**. Here, the peak of frequency counts in the third time interval, for the words "riot," "violence," "military," and "soldiers" is crystal clear.

## Discussion

In previous research on the WAR framing during the COVID-19 pandemic, it has been observed that roughly during the peak of the first wave of infections, between 20.03.20 and 02.04.20, about 5.32% of all tweets contained war-related terms (Wicke and Bolognesi, 2020). For a much longer timespan (20.03.20–01.07.20), we now report a similar trend: 5.60% of the tweets contain war-related terms. However, the number of war-related terms are not distributed equally in the corpus. A broad division of our data into three theoretically motivated timeframes (described in **Table 1**) shows that the total amount of war-related words is significantly different across time intervals. In particular, the total amount of war-related terms shows its highest concentration in the first interval, which we identified with the first global rise of infections. In this interval, the WAR frame worked particularly well-because the identification of the enemy was clear and unambiguous. The sense of risk and urgency toward the situation was under everyone's eyes, and so was the fear and anxiety associated with it. These, according to Flusberg et al. (2017) are the mappings that make the WAR figurative

**TABLE 2** | Results of the 2-way ANOVA for three time intervals and two constructs.

|  | dof | F | p | np2 |
|---|---|---|---|---|
| Constructs | 1 | 1616.590 | <0.001 | 0.892 |
| Time intervals | 2 | 106.427 | <0.001 | 0.523 |
| Constructs x time intervals | 2 | 180.998 | <0.001 | 0.651 |

*Dof, Degrees of Freedom. p-value: Uncorrected p-values. np2: Partial eta-squared effect sizes.*

**TABLE 3** | Results of the *post-hoc t*-test between construct and time intervals.

| Contrast | A | B | mean(A) | std(A) | mean(B) | std(B) | T-statistic | P* | Cohen's D |
|---|---|---|---|---|---|---|---|---|---|
| Construct | EventRelated | MostCommon | 41.080 | 31.213 | 291.580 | 118.737 | −20.404 | <0.001 | −2.886 |
| Intervals | Interval 1 | Interval 2 | 235.000 | 219.781 | 146.117 | 120.082 | 2.749 | 0.021 | 0.502 |
| Intervals | Interval 1 | Interval 3 | 235.000 | 219.781 | 129.988 | 80.814 | 3.527 | 0.002 | 0.672 |
| **Intervals** | **Interval 2** | **Interval 3** | **146.117** | **120.082** | **129.988** | **80.814** | **0.899** | **1.000** | **0.162** |
| EventRelated | Interval 1 | Interval 2 | 24.100 | 6.467 | 31.067 | 8.090 | −3.684 | 0.003 | −0.951 |
| EventRelated | Interval 1 | Interval 3 | 24.100 | 6.467 | 61.325 | 40.915 | −5.661 | <0.001 | −1.190 |
| EventRelated | Interval 2 | Interval 3 | 31.067 | 8.090 | 61.325 | 40.915 | −4.560 | <0.001 | −0.963 |
| MostCommon | Interval 1 | Interval 2 | 445.900 | 78.776 | 261.167 | 43.418 | 11.249 | <0.001 | 2.904 |
| MostCommon | Interval 1 | Interval 3 | 445.900 | 78.776 | 198.650 | 43.408 | 15.515 | <0.001 | 4.050 |
| MostCommon | Interval 2 | Interval 3 | 261.167 | 43.418 | 198.650 | 43.408 | 5.962 | <0.001 | 1.440 |

*Mean, Average value of occurrences over time; Std, Standard deviation of occurrences over time; p*, Bonferroni corrected p-value; Cohen's D, Effect size. In bold: row with no statistical significance.*

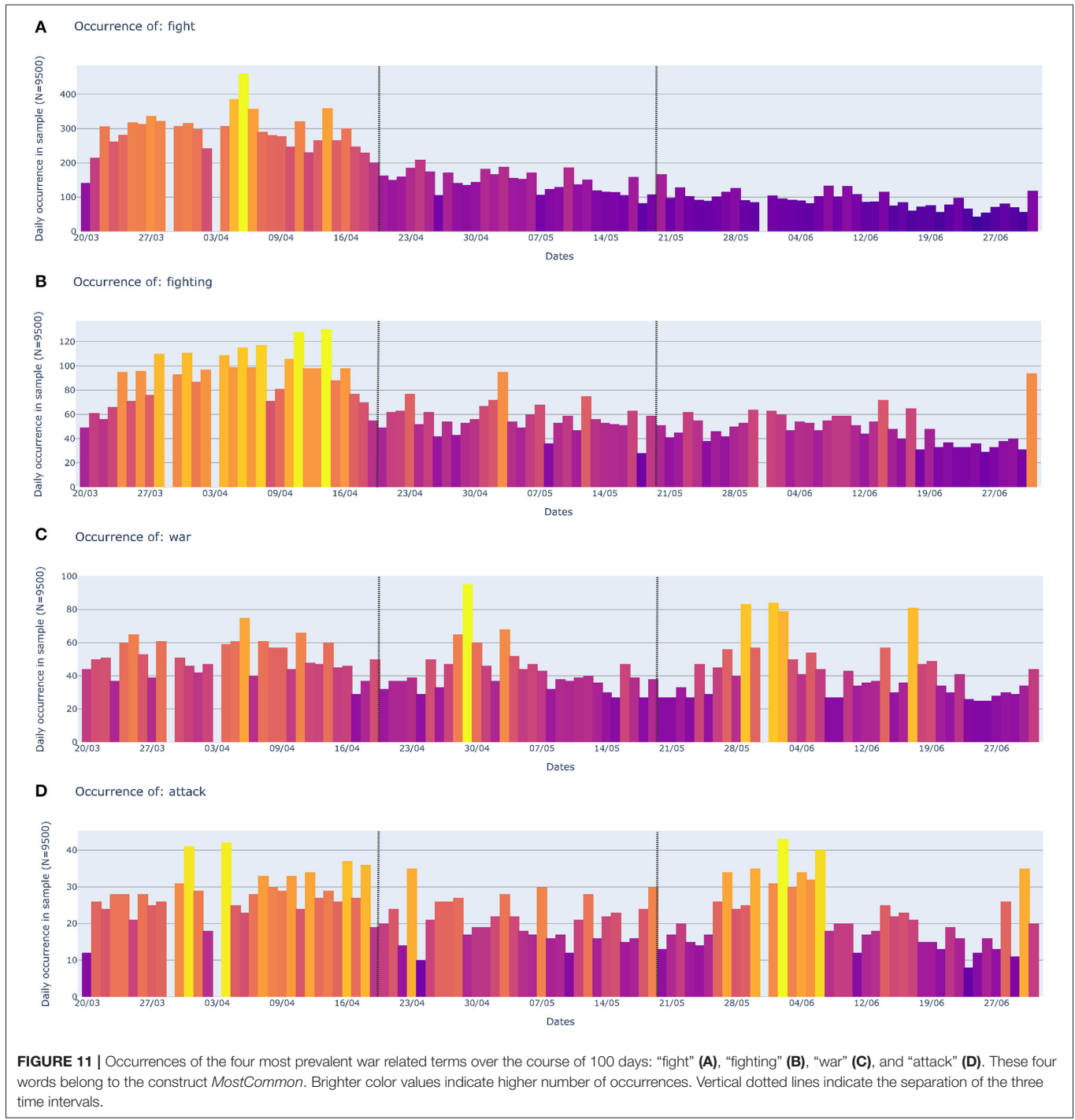

**FIGURE 11** | Occurrences of the four most prevalent war related terms over the course of 100 days: "fight" **(A)**, "fighting" **(B)**, "war" **(C)**, and "attack" **(D)**. These four words belong to the construct *MostCommon*. Brighter color values indicate higher number of occurrences. Vertical dotted lines indicate the separation of the three time intervals.

frame salient and effective in many communicative contexts. The number of war-related terms decreases substantially in the second interval, where the first global relaxation of infections is observed. In this situation, the war frame loses its salience and its efficacy. Finally, in the third interval the number of war-related terms is again significantly lower than in the previous interval. This interval, however, corresponds to the second global rise of infections with the highest number of daily new infections since the start of the pandemic. It is therefore quite interesting to observe that during this period the WAR frame did not rise again, as a second battle against the (same) enemy. This can be interpreted in different ways. One interpretation could consider that the WAR frame was not a very good fit anymore in the collective mind. In this sense, it is possible that alternative figurative frames may be preferred. In another interpretation, it could be the case that figuration in general might be not a good fit anymore, and that people prefer to use literal language. We hereby provide a possible explanation for this.

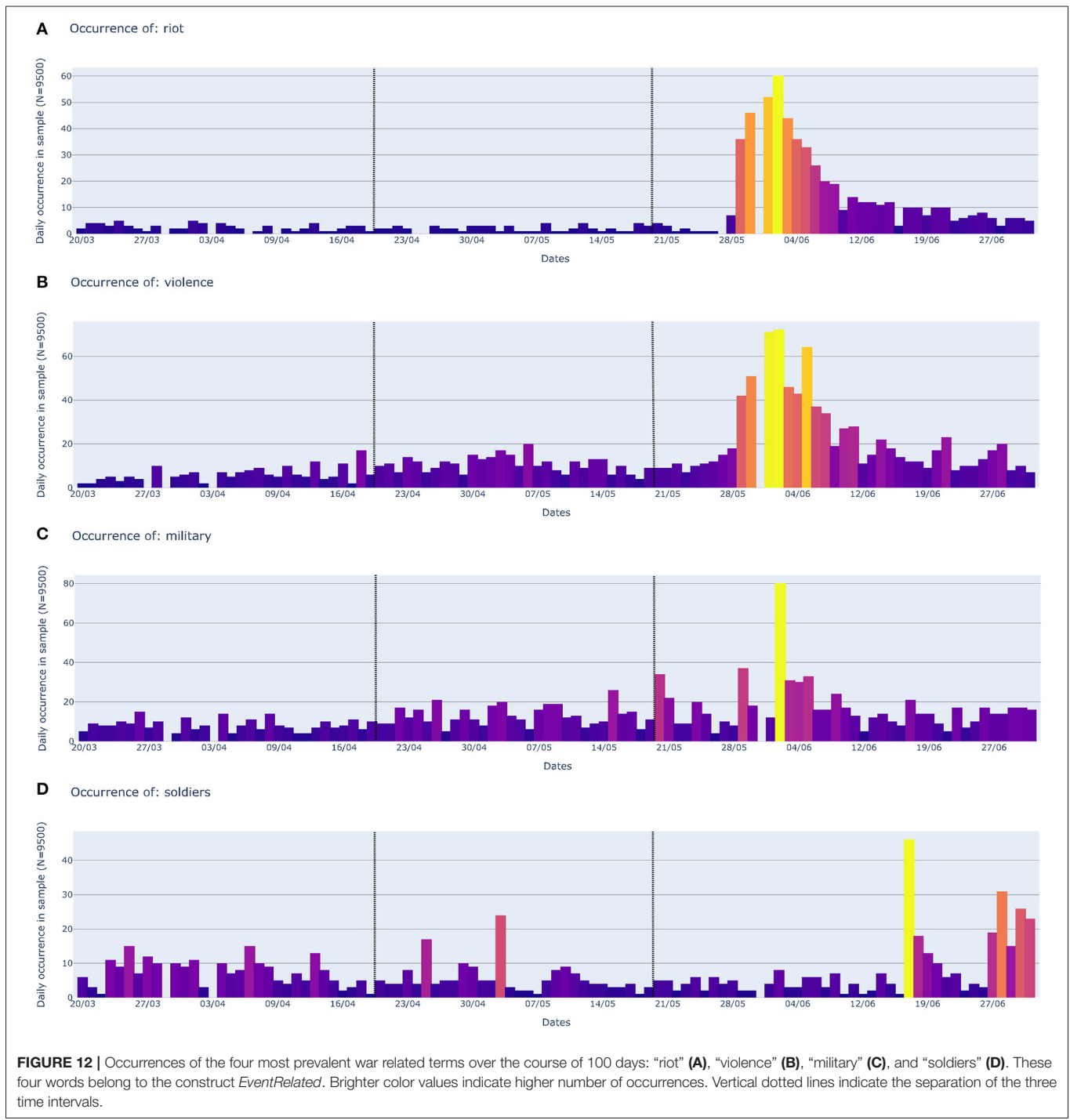

**FIGURE 12 |** Occurrences of the four most prevalent war related terms over the course of 100 days: "riot" **(A)**, "violence" **(B)**, "military" **(C)**, and "soldiers" **(D)**. These four words belong to the construct *EventRelated*. Brighter color values indicate higher number of occurrences. Vertical dotted lines indicate the separation of the three time intervals.

Looking at our data we can see a peak in war-related terms, roughly between 25th May and 15th June 2020. Informed by our topic modeling analysis, we can relate the murder of George Floyd and the BLM protests to this increase of war-related terms in the corpus. Notably, while the occurrence of words within the *MostCommon* construct ("fight," "fighting," and "war") decrease over time words within the *EventRelated* construct (end of May to mid-June) increase dramatically. These words are: "riot," "violence," "military," and "soldiers," as displayed in **Figure 12**. Crucially, these words are *not* used metaphorically. They are used in their literal meaning, in relation to the BLM protests sparked by the murder of George Floyd. In the third interval of our timeline, therefore, literal riots, soldiers, and violence have conflated with the figurative war-terms previously used to discuss the pandemic. The following examples show how the words "riot" and "violence," which are featured within the list of lexical entries for the WAR frame, are used in their literal meaning, in association to Covid-19:

> Covid was the setup *for the riots. Business closed...pre-covid to run out of business..post-covid so no one is around when rioting. Funny strange too the masks requirement in these cities giving cover...it's a warped strategy by Left/Dems to win election & or coup" 2nd July 2020

> "Think I need to switch off Twitter for a while, my feed is just full of senseless violence. #BlackLivesMatter for sure, but the world has temporarily forgotten we are also still fighting a global epidemic #Covid19" 2nd July 2020

Given the cruelty, vividness and reality of the literal events, it could be the case that figurative uses of war-terms lost (at least temporarily) their efficacy. These are open empirical questions that shall be investigated in the near future.

## GENERAL DISCUSSION AND CONCLUSION

In this paper we investigated how the topics, the emotional valence (sentiment polarity), the subjectivity, and the usage of words related to the WAR frame change over time, in the discourse about Covid-19.

We reported a series of qualitative and quantitative observations in relation to the changes of topics, based on topic models that partitioned our corpus into different numbers of clusters. We showed that more fine-grained solutions are likely to capture themes that do not emerge when the analysis is performed on a less granular scale. This is quite interesting, because by reporting observations on models that vary in their degree of granularity, we were arguably able to capture events that were reported in international but also national press. The implications of our findings may be relevant and informative for other studies that adopt topic modeling to explore large corpora of data. As a matter of fact, many studies tend to report analyses based on a unique topic modeling solution, that typically encompass a small number of topics (e.g., Lazard et al., 2015; Tran and Lee, 2016; Miller et al., 2017; Vijaykumar et al., 2017). These might miss crucial information that can be captured by a more fine-grained topic model. In our case more fine-grained analyses captured changes in the discourse and the entrance of new topics associated with Covid, such as the current protests that are taking place in the USA and see police and civilians involved in riots ignited by protest against social injustice and racial discrimination.

The analysis of the polarity of the sentiments expressed and the change in subjectivity through time has shown that the current pandemic affected our thoughts in different ways, during the timeframe considered in this study (and likely still does so, today). On one hand it pushed us toward deeper introspections and expression of beliefs and opinions, and on the other hand it induced negative feelings. The increase in subjectivity observed, seems particularly worrisome, because it could be interpreted as a gradual loss of trust in the data, in the facts, and in the objective information that should be conveyed by the press. The media may then be perceived to be increasingly less trustworthy. This trend, however, may be characterized by the medium we used to analyze our data: the social media platform Twitter. Recently, it has been shown that fake news spread six times faster than real news, on Twitter (Vosoughi et al., 2018). This sense of helplessness and skepticism toward the news found on Twitter is therefore substantiated, and it could also explain the negative feelings that characterize increasingly more often the tweets about Covid on this same platform. While these observations may remain rather exploratory, we suggest the following as a potential further investigation: does the increasing subjectivity and the increasing negative polarity of individuals' tweets correlate with a gradual loss of trust in the news that they read on this platform?

Finally, we reported a diachronic change in the way the WAR frame is used to talk about Covid. While the most frequent words within this frame, namely "fight," "fighting," and "war," and "attack" remain the most frequent words used in the Covid discourse, overall, their relative frequency of occurrence seems to decrease, concurrently with the overall decrease of the WAR framing through the progression of the pandemic, within the timeframe analyzed. We suggest that this could indicate that the WAR frame, used to figuratively frame the pandemic, may not always be a good fit, for all the stages of its development. Moreover, new words within the WAR frame become frequently used in the third time interval analyzed, in which we observed the emergence of topics associated with the riots in the USA that followed the murder of George Floyd. The lexical entries within the WAR frame that peak in this interval are different from those that peak in other intervals. In the third interval war-related words are used in their literal sense. "Riots," "violence," "military," and "soldiers" are real. They characterize actual events reported in the media, and are frequently used in tweets posted during this third time interval, in the corpus of Covid tweets. While further empirical investigation may be needed to support our interpretation, we suggest that because the discourse about Covid is now characterized by the description of events for which words within the WAR frame are used in their literal sense, then this could contribute to the overall decrease of metaphorical uses of war-related terms.

## DATA AVAILABILITY STATEMENT

The data for this study can be found in the Open Science Framework (OSF) at https://osf.io/v523j/?view_only= 63f03e24d48c4d58af1793e0f04ce28b.

## ETHICS STATEMENT

Written informed consent was not obtained from the individual(s) for the publication of any potentially identifiable images or data included in this article.

## AUTHOR CONTRIBUTIONS

PW and MB contributed to conception and design of the study. PW collected the data, organized the database and performed the statistical analysis. MB is responsible for writing sections introduction, theoretical background and related work, results in which topics are discussed on Twitter in relation to Covid-19 and how do they change over time with the development of the pandemic?, discussion in which topics are discussed on Twitter in relation to Covid-19 and how do they change over time with the development of the pandemic? discussion in what valence emerges from the tweets about Covid-19 and how does it change over time? discussion in how does the subjectivity of the tweets (i.e., opinion-based focus, vs. the objective fact-based focus) change over time?, discussion in how does the figurative framing of WAR change over time? PW is responsible for writing sections methods in which topics are discussed on Twitter in relation to Covid-19 and how do they change over time with the development of the pandemic?, Methods in what valence emerges from the tweets about Covid-19 and how does it change over time?. Results in what valence emerges from the tweets about Covid-19 and how does it change over time?, Methods in how does the subjectivity of the tweets (i.e., opinion-based focus, vs. the objective fact-based focus) change over time?, Results in how does the subjectivity of the tweets (i.e., opinion-based focus, vs. the objective fact-based focus) change over time?, Methods in how does the figurative framing of WAR change over time?, Results in how does the figurative framing of WAR change over time?. Results in Which topics are discussed on Twitter in relation to Covid-19 and how do they change over time with the development of the pandemic? and general discussion and conclusion were written together by the authors, who both revised and approved the submitted version. All authors contributed to the article and approved the submitted version.

## ACKNOWLEDGMENTS

The authors would like to thank Johannes Wiesner and Christian Burgers for their advice on statistical analysis.

## REFERENCES

Abd-Alrazaq, A., Alhuwail, D., Househ, M., Hamdi, M., and Shah, Z. (2020). Top concerns of tweeters during the COVID-19 pandemic: infoveillance study. *J. Med. Int. Res.* 22:e19016. doi: 10.2196/19016

Alves, A. L. F., de Souza Baptista, C., Firmino, A. A., de Oliveira, M. G., and de Figueirêdo, H. F. (2014). "Temporal analysis of sentiment in tweets: a case study with FIFA confederations cup in Brazil," in *International Conference on Database and Expert Systems Applications* (Springer, Cham), 81–88. doi: 10.1007/978-3-319-10073-9_7

Blei, D. M., Ng, A. Y., and Jordan, M. I. (2003). Latent dirichlet allocation. *J. Mach. Learn. Res.* 3, 993–1022. doi: 10.5555/944919.944937

Bruns, A., and Weller, K. (2016). "Twitter as a first draft of the present: and the challenges of preserving it for the future," in *Proceedings of the 8th ACM Conference on Web Science* (Hannover), 183–189. doi: 10.1145/2908131.2908174

Burgers, C., Konijn, E. A., and Steen, G. J. (2016). Figurative framing: shaping public discourse through metaphor, hyperbole, and irony. *Commun. Theory* 26, 410–430. doi: 10.1111/comt.12096

Chakraborty, K., Bhatia, S., Bhattacharyya, S., Platos, J., Bag, R., and Hassanien, A. E. (2020). Sentiment Analysis of COVID-19 tweets by deep learning classifiers—a study to show how popularity is affecting accuracy in social media. *Appl. Soft Comput.* 97:106754. doi: 10.1016/j.asoc.2020.106754

Chatterjee, S., Deng, S., Liu, J., Shan, R., and Jiao, W. (2018). Classifying facts and opinions in Twitter messages: a deep learning-based approach. *J. Business Analyt.* 1, 29–39. doi: 10.1080/2573234X.2018.1506687

Chaturvedi, I., Cambria, E., Welsch, R. E., and Herrera, F. (2018). Distinguishing between facts and opinions for sentiment analysis: survey and challenges. *Informat. Fusion* 44, 65–77. doi: 10.1016/j.inffus.2017.12.006

Culotta, A. (2010). "Towards detecting influenza epidemics by analyzing Twitter messages," in *Proceedings of the First Workshop on Social Media Analytics* (Washington DC: District of Columbia, USA), 115–122. doi: 10.1145/1964858.1964874

Damstra, A., and Vliegenthart, R. (2018). (Un) covering the economic crisis? Over-time and inter-media differences in salience and framing. *Journalism Studies*, 19, 983–1003. doi: 10.1080/1461670X.2016.1246377

De Smedt, T., and Daelemans, W. (2012). Pattern for python. *J. Mach. Learn. Res.* 13, 2063–2067. doi: 10.5555/2188385.2343710

Entman, R. (1993). Framing: toward clarification of a fractured paradigm. *J. Commun.* 43, 51–58. doi: 10.1111/j.1460-2466.1993.tb01304.x

Flusberg, S. J., Matlock, T., and Thibodeau, P. H. (2017). Metaphors for the war (or race) against climate change. *Environ. Commun.* 11, 769–783. doi: 10.1080/17524032.2017.1289111

Ghosh, A., and Veale, T. (2016). "Fracking sarcasm using neural network," in *Proceedings of the 7th Workshop on Computational Approaches to Subjectivity, Sentiment and Social Media Analysis* (San Diego, CA). 161–169. doi: 10.18653/v1/W16-0425

Gomide, J., Veloso, A., Meira Jr, W., Almeida, V., Benevenuto, F., Ferraz, F., et al. (2011). "Dengue surveillance based on a computational model of spatio-temporal locality of Twitter," in *Proceedings of the 3rd International Web Science Conference* (Koblenz). 1–8. doi: 10.1145/2527031.2527049

Hawkins, J. B., Brownstein, J. S., Tuli, G., Runels, T., Broecker, K., Nsoesie, E. O., et al. (2016). Measuring patient-perceived quality of care in US hospitals using Twitter. *BMJ Quality Safety* 25, 404–413. doi: 10.1136/bmjqs-2015-004309

Hendricks, R. K., Demján, Z., Semino, E., and Boroditsky, L. (2018). Emotional implications of metaphor: consequences of metaphor framing for mindset about cancer. *Metaphor. Symbol* 33, 267–279. doi: 10.1080/10926488.2018.1549835

Hu, T., She, B., Duan, L., Yue, H., and Clunis, J. (2019). A systematic spatial and temporal sentiment analysis on geo-tweets. *IEEE Access* 8, 8658–8667. doi: 10.1109/ACCESS.2019.2961100

Lamsal, R. (2020). *Data from: Coronavirus (COVID-19) Tweets Dataset.* IEEE Dataport. Available online at: http://dx.doi.org/10.21227/781w-ef42

Lazard, A. J., Scheinfeld, E., Bernhardt, J. M., Wilcox, G. B., and Suran, M. (2015). Detecting themes of public concern: a text mining analysis of the Centers for Disease Control and Prevention's Ebola live Twitter chat. *Am. J. Infect. Control* 43, 1109–1111. doi: 10.1016/j.ajic.2015.05.025

Li, I., Li, Y., Li, T., Alvarez-Napagao, S., and Garcia, D. (2020). *What are We Depressed About When We Talk About Covid19: Mental Health Analysis on Tweets Using Natural Language Processing.* arXiv preprint arXiv:2004.10899. Available online at: https://arxiv.org/abs/2004.10899 (accessed at: January 9, 2021). doi: 10.1007/978-3-030-63799-6_27

Li, R., Lei, K. H., Khadiwala, R., and Chang, K. C. C. (2012). "Tedas: a twitter-based event detection and analysis system," in *2012 IEEE 28th International Conference on Data Engineering.* 1273–1276. Arlington, VA: IEEE. doi: 10.1109/ICDE.2012.125

Liu, B. (2010). "Sentiment analysis and subjectivity," in *Handbook of Natural Language Processing Vol. 2*, eds N. Indurkhya, and F. J. Damerau (New York, NY: Chapman and Hall/CRC), 627–666. doi: 10.1201/9781420085938

Loria, S., Keen, P., Honnibal, M., Yankovsky, R., Karesh, D., and Dempsey, E. (2014). Textblob: simplified text processing. *Secondary TextBlob: simplified text processing*, 3.

Lwin, M. O., Lu, J., Sheldenkar, A., Schulz, P. J., Shin, W., Gupta, R., et al. (2020). Global sentiments surrounding the COVID-19 pandemic on Twitter: analysis of Twitter trends. *JMIR Public Health Surveillance* 6:e19447. doi: 10.2196/19447

Martín, Y., Li, Z., and Cutter, S. L. (2017). Leveraging Twitter to gauge evacuation compliance: spatiotemporal analysis of Hurricane Matthew. *PLoS ONE* 12:e0181701. doi: 10.1371/journal.pone.0181701

Mathur, A., Kubde, P., and Vaidya, S. (2020). "Emotional Analysis using Twitter Data during Pandemic Situation: COVID-19," in *2020 5th International Conference on Communication and Electronics Systems (ICCES).* Coimbatore: IEEE, 845–848. doi: 10.1109/ICCES48766.2020.9138079

Miller, M., Banerjee, T., Muppalla, R., Romine, W., and Sheth, A. (2017). What are people tweeting about Zika? An exploratory study concerning its symptoms, treatment, transmission, and prevention. *JMIR Public Health Surveillance* 3:e38. doi: 10.2196/publichealth.7157

Mollema, L., Harmsen, I. A., Broekhuizen, E., Clijnk, R., De Melker, H., Paulussen, T., et al. (2015). Disease detection or public opinion reflection? Content analysis of tweets, other social media, and online newspapers during the measles outbreak in The Netherlands in 2013. *J Med Int Res.* 17:e128. doi: 10.2196/jmir.3863

Park, S., Han, S., Kim, J., Molaie, M. M., Vu, H. D., Singh, K., et al. (2020). *Risk Communication in Asian Countries: COVID-19 Discourse on Twitter.* arXiv preprint arXiv:2006.12218. Available online at: https://arxiv.org/abs/2006.12218 (accessed at: January 9, 2021).

Paul, D., Li, F., Teja, M. K., Yu, X., and Frost, R. (2017). "Compass: Spatio temporal sentiment analysis of US election what twitter says!," in *Proceedings of the 23rd ACM SIGKDD International Conference on Knowledge Discovery and Data Mining* (Halifax, NS). 1585–1594. doi: 10.1145/3097983.3098053

Pruss, D., Fujinuma, Y., Daughton, A. R., Paul, M. J., Arnot, B., Albers Szafir, D., et al. (2019). Zika discourse in the Americas: a multilingual topic analysis of Twitter. *PLoS ONE* 14:e0216922. doi: 10.1371/journal.pone.0216922

Rehurek, R., and Sojka, P. (2010). Software framework for topic modeling with large corpora. In In Proceedings of the LREC 2010 Workshop on New Challenges for NLP Frameworks.

Reyes, A., Rosso, P., and Veale, T. (2013). A multidimensional approach for detecting irony in twitter. *Language Resources Evaluat.* 47, 239–268. doi: 10.1007/s10579-012-9196-x

Reynard, D., and Shirgaokar, M. (2019). Harnessing the power of machine learning: can Twitter data be useful in guiding resource allocation decisions during a natural disaster? *Transportation Res. Part D* 77, 449–463. doi: 10.1016/j.trd.2019.03.002

Semino, E., Demjén, Z., Demmen, J., Koller, V., Payne, S., Hardie, A., et al. (2017). The online use of Violence and Journey metaphors by patients with cancer, as compared with health professionals: a mixed methods study. *BMJ Support. Palliative Care* 7, 60–66. doi: 10.1136/bmjspcare-2014-000785

Sievert, C., and Shirley, K. (2014). "LDAvis: a method for visualizing and interpreting topics," in *Proceedings of the Workshop on Interactive Language Learning, Visualization, and Interfaces* (Baltimore, MD). 63–70. doi: 10.3115/v1/W14-3110

Sontag, S., and Broun, H. H. (1977). *Illness as Metaphor (p. 343).* Farrar, Straus.

Statista (2020). Available online at: https://www.statista.com/ (accessed at: February 10, 2021).

Stone, B., Dennis, S., and Kwantes, P. J. (2011). Comparing methods for single paragraph similarity analysis. *Top. Cognit. Sci.* 3, 92–122. doi: 10.1111/j.1756-8765.2010.01108.x

Syed, S., and Spruit, M. (2017). "Full-text or abstract? Examining topic coherence scores using latent dirichlet allocation," in *2017 IEEE International Conference on Data Science and Advanced Analytics (DSAA).* IEEE. 165–174. doi: 10.1109/DSAA.2017.61

Thibodeau, P. H., Hendricks, R. K., and Boroditsky, L. (2017). How linguistic metaphor scaffolds reasoning. *Trends Cognit. Sci.* 21, 852–863. doi: 10.1016/j.tics.2017.07.001

Tran, T., and Lee, K. (2016). "Understanding citizen reactions and Ebola-related information propagation on social media," in *2016 IEEE/ACM International Conference on Advances in Social Networks Analysis and Mining (ASONAM).* IEEE. 106–111. doi: 10.1109/ASONAM.2016.7752221

Vijaykumar, S., Nowak, G., Himelboim, I., and Jin, Y. (2017). Virtual Zika transmission after the first U.S. case: who said what and how it spread on Twitter. *Am. J. Infect. Control* 46, 549–557. doi: 10.1016/j.ajic.2017.10.015

Vosoughi, S., Roy, D., and Aral, S. (2018). The spread of true and false news online. *Science* 359, 1146–1151. doi: 10.1126/science.aap9559

Wei, G., Lim, E. P., and Zhu, F. (2015). *Characterizing Silent Users in Social Media Communities.*

Wicke, P., and Bolognesi, M. M. (2020). Framing COVID-19: how we conceptualize and discuss the pandemic on Twitter. *PLoS ONE* 15:e0240010. doi: 10.1371/journal.pone.0240010

Wirz, C. D., Xenos, M. A., Brossard, D., Scheufele, D., Chung, J. H., and Massarani, L. (2018). Rethinking social amplification of risk: Social media and Zika in three languages. *Risk Analysis* 38, 2599–2624. doi: 10.1111/risa.13228

World Health Organization (2020). Available online at: https://www.who.int/emergencies/diseases/novel-coronavirus-2019 (accessed at: February 10, 2021).

Xue, J., Chen, J., Hu, R., Chen, C., Zheng, C., Su, Y., et al. (2020). Twitter discussions and emotions about the COVID-19 pandemic: machine learning approach. *J. Med. Int. Res.* 22:e20550. doi: 10.2196/20550

Yeo, J., Knox, C. C., and Hu, Q. (2020). *Disaster Recovery Communication in the Digital Era: Social Media and the 2016 Southern Louisiana Flood. Risk Analysis.* Available online at: https://doi.org/10.1111/risa.13652 doi: 10.1111/risa.13652

Yeo, J., Knox, C. C., and Jung, K. (2018). Unveiling cultures in emergency response communication networks on social media: Following the 2016 Louisiana floods. *Quality Quantity* 52, 519–535. doi: 10.1007/s11135-017-0595-3

Zhou, J., Yang, S., Xiao, C., and Chen, F. (2020). *Examination of Community Sentiment Dynamics Due to Covid-19 Pandemic: A Case Study From Australia.* arXiv preprint arXiv:2006.12185. Available online at: https://arxiv.org/abs/2006.12185 (accessed at: January 9, 2021).

**Conflict of Interest:** The authors declare that the research was conducted in the absence of any commercial or financial relationships that could be construed as a potential conflict of interest.